\documentclass{article}

\usepackage{PRIMEarxiv}

\usepackage[utf8]{inputenc} 
\usepackage[T1]{fontenc}    
\usepackage{hyperref}       
\usepackage{url}            
\usepackage{booktabs}       
\usepackage{amsfonts}       
\usepackage{nicefrac}       
\usepackage{microtype}      
\usepackage{lipsum}
\usepackage{fancyhdr}       
\usepackage{graphicx}       
\graphicspath{{media/}}     

\usepackage{multirow}
\usepackage{comment}
\usepackage{subfigure}
\usepackage{graphicx}
\usepackage{rotating}
\usepackage{hyperref}

\title{Understanding Activation Patterns in Artificial Neural Networks by Exploring Stochastic Processes: Discriminating Generalization from Memorization }

    \author{
  Stephan Johann Lehmler \\
    Institute of Computer Science\\
Ruhr West  University of Applied Science \\
  Mülheim an der Ruhr, Germany,\\
  Faculty of Electrical Engineering and \\Information Technology \\
  Ruhr University Bochum,   Bochum, Germany\\
  \texttt{lehmler@med-psych.uni-kiel.de} \\
   \And
  Muhammad Saif-ur-Rehman  \\
  Institute of Computer Science\\
Ruhr West  University of Applied Science \\  
  Mülheim an der Ruhr, Germany\\
  \texttt{Muhammad.Saif-ur-Rehmann@hs-ruhrwest.de}
  \And
   Tobias Glasmachers  \\
  Institute for Neural Computation\\
  Ruhr University Bochum\\
  Bochum, Germany\\
  \texttt{tobias.glasmachers@ini.rub.de}
     \And
Ioannis Iossifidis  \\
  Institute of Computer Science\\
Ruhr West  University of Applied Science \\
  Mülheim an der Ruhr, Germany\\
  \texttt{iossifidis@hs-ruhrwest.de}
}

\begin{document}
\maketitle

\begin{abstract}
To gain a deeper understanding of the behavior and learning dynamics of (deep) artificial neural networks, it is valuable to employ mathematical abstractions and models. These tools provide a simplified perspective on network performance and facilitate systematic investigations through simulations. In this paper, we propose utilizing the framework of stochastic processes, which has been underutilized thus far.

Our approach involves modeling the activation patterns of thresholded nodes in (deep) artificial neural networks as stochastic processes. By disregarding the magnitude of activations and focusing solely on the activation frequency, we can leverage techniques commonly used in neuroscience to study spike trains observed in real neurons. Specifically, we extract the spiking activity of individual artificial neurons/nodes during a classification task and model their spiking frequency. The underlying process model we employ is an arrival process that follows the Poisson distribution.

We examine the theoretical fit of the observed data generated by various artificial neural networks in image recognition tasks to the proposed model's key assumptions. Through the stochastic process model, we derive a set of parameters that describe the activation patterns of each network. Our analysis encompasses randomly initialized networks, generalizing networks, and memorizing networks, allowing us to identify consistent differences across multiple architectures and training sets.

We calculate several features, including Mean Firing Rate, Mean Fano Factor, and Variances, which prove to be stable indicators of memorization during learning. These calculated features offer valuable insights into network behavior. Moreover, the proposed model demonstrates promising results in describing activation patterns and could serve as a general framework for future investigations. It has potential applications in theoretical simulation studies as well as practical areas such as pruning or transfer learning.

\end{abstract}

\keywords{Artificial Neural Networks \and Stochastic Modeling \and Poisson Process \and Generalization \and Memorization}


\section{Introduction}
While deep learning continues to solve many real world problem, there has been no shortage of mathematical and empirical approaches to better understand what are indications of successful learning (e.g., \cite{berner_modern_2021,jin_quantifying_2020,jin_how_2020,laakom_within-layer_2021,neyshabur_exploring_2017, zhang_understanding_2017,roberts_principles_2022}). The activation patterns of an artificial neural network (ANN) while processing given inputs has been the focus of much of the existing empirically oriented research (e.g., \cite{laakom_within-layer_2021,gain_abstraction_2020,liu_neuron_2023,jin_quantifying_2020,banerjee_empirical_2019,guiroy_improving_2022} . 

Measuring and examining activation patterns of nodes within an ANN seems to be a necessary step forward towards understanding what features of a network correlates with generalization. It seems obvious that such a relationship between the activations within an ANN and its generalization performance exist. However, no commonly accepted framework for investigating this relationship exists yet.\\
An active research area to understand the relation activation pattern and learning dynamics has been feature visualization of activations within a layer, specific nodes or with regard to input classes \cite{nguyen2016multifaceted, yosinski2015understanding, selvaraju2017grad, nguyen2016synthesizing}. Related works, such as saliency maps \cite{simonyan2014deep, adebayo2018sanity}, sensitivity analysis \cite{pizarroso2020neuralsens} or adversarial attacks \cite{nguyen2015deep} investigate how activations rely on specific inputs. While these works offer exciting new ways to look at what happens inside an ANN,however,  it stays difficult to extract quantitative measures that are capable of differentiating between memorization and generalization in neural network learning.\\
Activations, especially their magnitudes, distributions, or gradients play also an important role in pruning\cite{blalock2020state, ye2020accelerating, anwar2017structured, zhao2022adaptive, hu2016network, tan2020dropnet} and regularization (e.g., \cite{ding2019regularizing, joo2020regularizing, qi2019activity, hanin2019deep, merity2017revisiting}). Pruning and regularization both proved immensely important not only for practical applicability of ANNs but also for theoretical understanding of generalization. The rules derive however are often heuristics without a unified view on how activations are contributing to generalization or memorization. Some derived heuristics, also appear to be contradictory, e.g., regularization is usually concerned with keeping activation magnitudes low \cite{merity2017revisiting}, while many pruning techniques are trying to remove nodes with low activations \cite{tan2020dropnet}. \\
\\
This paper looks at activations within a network, but offers a new perspective utilizing statistical modeling methods commonly found in neuroinformatics and computational neuroscience. We are presenting a simple mathematical formulation for analyzing networks in the context of memorization and generalization, that to the best of our knowledge hasn't been utilized anywhere in the literature so far.\\
Given an artificial neural network, we analyze how a number of chosen nodes (in this paper, we limit ourselves to the last hidden layer) of the ANN 'react' to known or unknown inputs. When feeding $n$ samples of a dataset through an ANN, the behavior of each observed neuron/node in the network can be seen in the form of a time series of length $n$. Given these vectors of activity (or non-activity), we want to investigate whether the observed activation frequency gives insight into the generalization ability of the network. As an intuitive example, we would expect an ANN that perfectly memorized its input to have many nodes that only activate once for one specific memorized sample.\\ 
We start by simplifying the analysis by discarding the magnitude of neural activations. Ignoring the magnitude of each activation does throw away some potentially important information, but the magnitude is also highly dependent of inputs and utilized activation function. By thresholding and binarizing the observed activations, we only keep a binary vector of size $n$, telling us whether the given node did activate or not when seeing a given input. We hope this approach keeps significant information about activation patterns, while being flexible and more independent of used activation functions or input scaling. \\
The key advantage of this simplified approach is that these binarized vectors of observed activations can be interpreted as a neural spike train and as such could be analyzed with techniques commonly utilized within neuroinformatics and computational neuroscience.
For our analysis, we model the observed spike train stochastically as an arrival process following a Poisson distribution.Poisson distributions are commonly encountered as baseline models in various scientific disciplines, including computational neuroscience, as their assumptions fit most independent arrival processes.  Details of the modeling approach and assumptions will be given in the following chapter.\\
The same or related models are commonly used by neuroscientists for modeling the behavior of real neurons (e.g. \cite{shadlen1994noise, Softky334, deger2009poisson, reynaud2013spike, PhysRevE.73.022901, Kass2014, kramer2016case}). Computational neuroscience also used the framework of stochastic processes for modeling activation patterns of stochastic or pulsing neural networks (e.g. \cite{brown2001stochastic, 963787, 1058080, CARD2001173, cowan1990stochastic, card1998doubly, doi:10.1142/S0129065701000552,     YANG201887} but also more general network models (e.g. \cite{pregowska_signal_2021,  PhysRevA.44.2718, goltsev2010stochastic}). In engineering, especially signal processing, the connection between Poisson processes and fault-tolerant processing has been noted \cite{keane2001impulses, ma2012high}, but applications to neural networks are rare. In the context of Bayesian Radial Basis Function Networks, \cite{661204} used a Poisson process to derive activation priors and \cite{hanin2021random} show a mathematical connection between Gaussian processes and infinite width random neural networks. Stochastic processes have more often been used as an application for neural networks (e.g. \cite{PhysRevD.103.116023, Lee_Zame_Yoon_vanderSchaar_2018, NEURIPS2021_f19c44d0}) but have, to the best of our knowledge, not yet been used for analyzing generation capabilities of ANN.
The more general idea of firing frequency \cite{qi2019activity, hu2016network} and activation patterns 
\cite{hanin2019deep, joo2020regularizing} as important features related to generalization has been proposed before, but not yet modeled rigorously.\\
\\
We believe that successfully finding stable measures based on our stochastic framework would allow us to provide more meaningful insight about the important properties (generalization \& memorization) of employed architectures and their parameters without having to rely solely on the accuracy of a training set. The use of stochastic processes and related approaches might open up new ideas for pruning, regularization, as well deepen our understanding about what is going on inside the network during learning/generalization and memorization. Furthermore, we believe that knowing about the stochastic properties of  well-behaving could open up the use of stochastic simulation for deriving new hypothesis and finding new ways to look at ANNs training dynamics.\\

\section{Stochastic Modeling}
As stated above, we are going to model the observed thresholded activations of one neuron while processing a data set as a stochastic process. For this, we need to clarify some vocabulary and models common to stochastic modeling approaches. We refer to the books by \cite{cinlar2013introduction} and \cite{nelson2010stochastic} for a thorough introduction and discussion of stochastic process modeling.
\subsection{Stochastic processes}
A stochastic process $X(t)$ can be defined as a grouping of random variables $\{X_t, t\in T\}$, where each $X$ is a random variable defined over a sample space $\Omega$. The index set $T$  usually denotes the time that the random variable was measured. If $T$ is countable, the stochastic process is called a \textit{discrete time process}, if $T$ is not countable, we speak of a \textit{continuous time process}. In our application $T$ will denote the position in the data set that is shown to the ANN, we therefore only consider discrete time processes.
Another important property that define many stochastic processes is \textit{stationarity}, meaning that its average statistical properties are
independent of where they are formed along the time axis\cite{GABBIANI2017335}. A given statistical measure $f$ (e.g., mean or variance) of $X$  would not depend on  $T$, formally we define a \textit{stationary stochastic process} (with regard to $f$) as a stochastic process, where $f(X_1 , ..., X_n) =f(X_{1+r} , ..., X_{n+r})$ for all $r$. The models applied are assumed to be stationary. All models considered are furthermore \textit{ergodic} processes. A process is regarded as $f$-ergodic if the time average $f$ equals the theoretical ensemble averaged $f$. As an example, a process is mean-ergodic if $E[P] = \lim_{t\to\infty} [ \int_{0}^{\infty} { X_t\,dx }]$

\subsubsection{Arrival Process}
We will be working with a specific type of stochastic process, an \textit{arrival (counting) process} (page 71,  \cite{cinlar2013introduction}). The general arrival process is defined on the interval $ t = [0, \infty]$ where the mapping $ t \rightarrow X_t(\omega)$ results for every $\omega \in \Omega$ in discrete (integer) values for which hold $X_t \ge  0$ (non-negativity), if $s \le t$ then  $X_t \ge  X_s$ (non-decreasing) and  $X_0(\omega) = 0 $ (process starts at 0).
Such a process can be a useful model for counts of observed events in a system, e.g., arrivals in a queue or spikes produced by a neuron.  
\subsubsection{Poisson Process}
A Poisson Process is an arrival process with additional constraints on the type of increments/arrivals modeled. If arrivals happen (1) one-at-a-time and are (2) independent as well as (3) stationary, the arrival process is a Poisson process.   Any arrival process that fits these qualitative criteria must follow a Poisson distribution (page 97, \cite{nelson2010stochastic}  and page 72 \cite{cinlar2013introduction}). A Poisson process is mathematically described that for all $t\geq0$ , $P(X_t = n) = \frac{(\lambda t)^n}{n} e^{-\lambda t}$ for some real-valued constant $\lambda \geq 0 $ and some integer $n$.  The only parameter defining the Poisson distribution is the arrival rate $\lambda$ . The Poisson distribution gives the probability of having observed $n$ arrivals at time $t$.\\
The Poisson process described above by only one parameter is the stationary or homogeneous Poisson process.  Additional models exist to model inhomogeneous processes, where the arrival rate $\lambda$ is allowed to vary. 

Two interesting properties of Poisson processes, that give some insight into how multiple processes interact with each other, are the superposition and decomposition property. (p97-88 \cite{cinlar2013introduction} p99-104 \cite{nelson2010stochastic}) Given two \textit{independent} Poisson processes, $A$ with $\lambda_A$ and $B$ with $\lambda_B$, they can be composed additively to form a new Poisson process $C$, with $\lambda_c = \lambda_A + \lambda_B$ (superposition property). The related decomposition property is the inverse property, stating that a Poisson process can be decomposed subtractively into distinct processes. It should be stressed that both properties only hold for independent processes. 

\subsubsection{Related Distributions}
The Poisson process is related to and can be characterized by some other well-known distributions.
The eponymous Poisson distribution gives the probability of  having observed $n$ arrival after $t$ time steps of a Poisson process. For the same Poisson arrival process, we could however also be interested in the distribution of waiting time until one arrival occurs. This probability is modeled by the exponential distribution $ P(T = t) = \lambda e^{-\lambda t}$.  The arrival rate $\lambda$ is the same parameter as  for the Poisson distribution. We deal with two different but valid descriptions of the same process.\\
By describing the Poisson process with an exponential distribution in terms of waiting times, we are able to formulate another important property.  The \textit{memoryless} property states that the observation of one arrival does not affect the probability of the waiting time until the next arrival. \\ 
We also might be interested in the time it takes for a specific number $k$ of arrivals to take place. In this case, we model the Poisson process using the Erlang distribution $P(T=t) = \frac{\lambda(\lambda t)^{k-1}e^{-\lambda t}}{(n-1)!}$ . For $k=1$, this reduces to the exponential distribution. \\
More generally, a homogeneous Poisson process is essentially a sequence of Bernoulli trials. (\cite{Kass2014})  Any Poisson process is the summation of $t$ independent Bernoulli trials, each of which can be modeled by a binomial distribution $ P(k) = {n\choose k} p^kq^{n-k}$,  the Poisson distribution is well-known to be the limiting case of the binomial distribution.\\

\subsection{Modeling activation patterns}
Stochastic processes have long been applied in neuroscience to model and characterize the activation pattern of real-life neuron (e.g., \cite{Kass2014}\cite{maimon_beyond_2009}\cite{stella_comparing_2021}\cite{berry_structure_1997}). 
The basic approach consists of recording a time series of spiking behavior from a singular neuron or a group of neurons. One can either use these spike-train data directly to fit a Poisson distribution, or calculate the time between the arrivals of two spikes, the inter-spike intervals (ISI), and fit an exponential distribution. The basic advantage of using stochastic process models over just calculating the average firing rate is that a stochastic model gives us access to the theoretical instantaneous firing rate, while averaging give a noisy point estimate, that depend more strongly on factors such as windowing size (\cite{Kass2014}).\\
After estimating model parameters, possible applications are manifold, such as comparing brain areas \cite{maimon_beyond_2009}, as a basis to measure effects of stimuli \cite{ramezan_multiscale_2014} \cite{naud_improved_2011} or a surrogate models for hypothesis testing \cite{stella_comparing_2021} or as a model for the neural code \cite{berry_structure_1997} with applications such as modeling the input into more biophysically detailed models (\cite{williams_point_2020}).\\
The Poisson process is a popular model in neuroscience, for reasons of simplicity rather than fidelity to nature.  Besides arguments based on empirical observations \cite{PhysRevE.73.022901} \cite{reynaud2013spike}, there exist many issues with the core assumptions explained above. The non-linear dynamics of neural activations, such as the refractory period \cite{kramer2016case}, cause real neurons to violate the core assumptions behind a Poisson process. Increments tend to not be independent and the activations non-stationary.

\subsubsection{Activations as spike trains}
We expect that artificial neurons in deep neural networks can be a more suitable system to model as a Poisson process. Compared to real neurons, those simplified ANNs  do more closely follow the core properties of the Poisson process. \\
Simple artificial neural networks (ANNs) without recurrent connections are considered memoryless, meaning they lack the ability to retain information over time. These networks exhibit stationary firing rates, assuming random samples are drawn from a dataset. Furthermore, the activation of individual neurons or nodes in the network occurs sequentially and independently of one another, assuming random samples are drawn from a dataset.

The hypothetical caricature of a perfectly overfitted model would however violate the independence assumption, as the firing probability would go to zero after firing once.

\subsubsection{Process description}
For viewing an ANN as a producer of spike trains analogous to neurons, we observe a trained ANN while it processes a dataset. One processed sample would equal one time point, we therefore will only consider discrete time models.
After observing the activations of one node or a group of nodes and applying a thresholding function, binarizing every node with activation values above 0 to 1, we end up with a spike train of activations per node, which can be modeled as a Poisson Arrival Process. Analyzing the calculated  parameters of the stochastic model, or distributions thereof (in case we are looking at groups of nodes) for different networks or training settings,  helps us gain insight into how activation patterns reflect training  success.

 \begin{figure}[h!]
  \includegraphics[width=\textwidth]{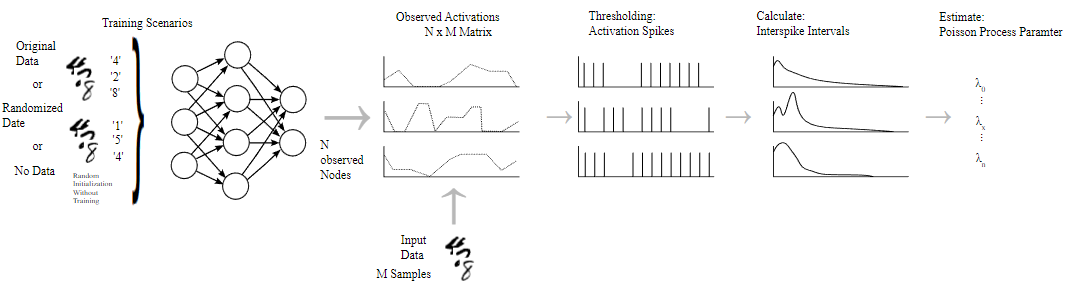}
  \caption{Analysis Steps. Starting from the randomly initialized, generalizing or memorizing network, we observe the last hidden layer while the network processes input data. The observed activations are thresholded to create binary activation spike trains, which are the used to calculate the firing rate/ $\lambda$-parameter of the assumed Poisson Process}
  \label{fig:process}
\end{figure}

\section{Analysis of ANN Activation Patterns}
We set out to test the applicability and usefulness of the Poisson Arrival Model to ANNs on a range of relatively simple but important image recognition benchmark data.  Our two main goals are (1) investigations into how suitable the assumptions and properties of the Poisson Process are to ANNs trained on real data and (2) evaluate whether it is possible to distinguish between generalizing and memorizing networks using simple indices in the Poisson process model. \\

\subsection{Method}
The overall approach closely follows the process description and figure \ref{fig:process}. The ANN models are first randomly initialized, then either trained on the original training data with the aim to generalize on the test set. Alternatively,  trained on the training data with randomized labels for memorizing the inputs (inspired by memorization tests popularized e.g. in \cite{zhang_understanding_2021}). For the memorizing networks, we configured and compared three different settings : (a) a retraining situation, where we start from the generalized networks weights, all layers remain frozen except the last two layers and then train on the shuffled data [MemRetrainLast] (b) the same retraining situation, without freezing any layers [MemRetrainAll] and (c) a  network with randomly initialized parameters (weights and biases). The parameters of all the layers remain frozen, except the last two layers  [MemRandomLast].

After training, to distinguish generalization and memorization, the resulting models are used to classify both training and test set. During this step, we observe the activation of nodes of one chosen layer. In principle, any or multiple layers could be observed for our experiments. However,  we restricted ourselves to the last hidden layer. For the sake of feasibility, we constrained this paper to one layer, and we selected this layer because some findings suggest that the latter layers in a network tend to be more related to memorization \cite{stephenson_geometry_2021,mo_towards_2019,arpit_closer_2017,cohen_dnn_2018,wongso_using_2023}. Differences between the chosen layers could be investigated in later studies.\\ 
The observed activations produced by each node for each sample of the dataset are thresholded to get activation spike trains. For each node in the observed layer, we therefore got a binary time series (t = number of samples), which can be modeled by a Poisson Arrival Model as discussed above in section 2.2. \\
Based on the statistics of these stochastic models, we analyzed how well the Poisson Model fits the data and how useful summary statistics are for distinguishing generalization and memorization.

\subsubsection{Datasets and Tasks}
As basis for our experiment, we use the common image recognition benchmark datasets MNIST, FashionMNIST and CIFAR-10. To aid reproducibility, we used the datasets as supplied by keras \cite{chollet_keras_2015} ),  using the train/test-split provided by the package. The only minimal preprocessing being the conversion of the integer labels to binary class vectors and normalizing the inputs (to bound the input values between 0 and 1).  The data was used to train the ANNs to classify the input images into one of ten classes. \\
We recorded the spiking activity as described above under five conditions of model training (random weights, trained for generalization and the three forms of memorization). For the random condition, we took the untrained models, for the generalization condition, we trained the models on the data with the aim to achieve high validation accuracy, and lastly for the memorization condition, we shuffled the input before training to enforce memorization. \\
We did focus on this comparatively simple set-up because these smaller datasets allow us to relatively easily find and train models capable of memorization. While future experiments covering a wider range of applications could be interesting,  we believe that the datasets chosen show enough variety for our experiments, and we don't have reason to assume, at this point, that different data sources would change the results drastically.

\subsubsection{Models and Parameters}
For each dataset, we chose models suitable to the task of achieving high classification accuracy on the respective task. The model and hyperparameter selection was done by grid-search, with the aim to find models that would be able to generalize well when trained on the original data and still be able to memorize the training data with random labels. The hyperparameter selection was done  independent of any spike-train analysis, to avoid biasing the results.
For the MNIST and FashionMNIST, we used the same set of models. The first model is a simple dense multilayer perceptron, with only one hidden layer after the flattened input and before the output layer [MLP]. The second model is a simple convolutional neural network [CNN], where the input was processed by a 2D-convolutional layer with 32 filters and a kernel size of 3. The outputs of the convolutional layer are flattened and afterward processed by the same architecture as with the dense model. We varied the size of the hidden dense layer and tested 64, 128, 256 and 512 nodes. The output was in both cases a 10-node dense-layer.  Activation function for each layer were rectified linear units.\\
For cifar10, we had to manually test a wider range of models and optimizers. The basic preprocessing architecture chosen was DenseNet121 [denseNet] as provided by keras \cite{huang_densely_2017}, with weights pretrained on IMAGENET and an input shape of (32,32,3). The outputs of this network were flattened and again processed by a dense ReLU-Layer of variable size. \\
Training of all the models was done using the same optimizer and hyperparameters. We trained using stochastic gradient descent with a learning rate of 0.1, momentum disabled and a batch size of 256  using a 10\% validation split.\\
The training and validation accuracies achieved can be seen in table \ref{table:acc}. All networks tested were able to generalize well on the uncorrupted dataset and all could memorize on the randomized data, albeit to varying degree depending on network depth and width and retraining method.

\begin{table}[htb]
\centering
\caption{\label{table:acc} Table showing the accuracy of each model achieved on the training and the validation set under different conditions. We see that all models reach a satisfactory accuracy in the generalization setting without a too large generalization gap. The memorizing networks manage to memorize parts of the shuffled training data, but the amount of samples memorized varies with data, model size and training condition. Due to the size of the MLP (only one hidden layer), MemRetrainAll is identical to MemRetrainLast and is therefore not reported.}
\resizebox{\textwidth}{!}{
\begin{tabular}{c|c|c|c|c|c|c|c}
\hline
\multirow{2}{*}{Model} & \multirow{2}{*}{Width} & \multirow{2}{*}{Data} & \multirow{2}{*}{Data split} & Generalizing & MemRetrainLast & MemRandomLast & MemRetrainAll \\
&&&&  Accuracy  & Accuracy & Accuracy & Accuracy \\
\hline
\multirow{16}{*}{MLP} & \multirow{4}{*}{64} & \multirow{2}{*}{MNIST}&\ Train & 98.66 & 31.11 & 41.22 & -\\
  &  &\ &\ Validation& 97.10  & 10.83 & 10.28& -\\
  &  & \multirow{2}{*}{FashionMNIST}&\ Train & 90.96 & 29.71 & 47.59 & -\\
  &  &\ &\ Validation& 88.85 & 10.67 & 09.02& -\\
& \multirow{4}{*}{128} & \multirow{2}{*}{MNIST}&\ Train & 99.03 & 58.84 & 63.80 & -\\
  &  &\ &\ Validation& 97.57 & 10.13 & 09.87& -\\
  & & \multirow{2}{*}{FashionMNIST}&\ Train & 91.75 & 55.43 & 74.88 & -\\
  &  &\ &\ Validation& 98.28 & 09.70 & 09.67& -\\
& \multirow{4}{*}{256} & \multirow{2}{*}{MNIST}&\ Train & 99.17 & 93.28 & 98.06 & -\\
  &  &\ &\ Validation& 97.62& 10.20 & 10.70& -\\
  &  & \multirow{2}{*}{FashionMNIST}&\ Train & 92.16 & 82.51 & 97.38 & -\\
  &  &\ &\ Validation& 89.27 & 09.47 & 09.98 & -\\
  & \multirow{4}{*}{512} & \multirow{2}{*}{MNIST}&\ Train & 99.34 & 100 & 100 & -\\
  &  &\ &\ Validation& 97.95 & 10.20 & 10.55& -\\
  &  & \multirow{2}{*}{FashionMNIST}&\ Train & 92.64 & 98.84 & 99.99 & -\\
  &  &\ &\ Validation& 89.63 & 09.58 & 10.53& -\\
  \hline

\multirow{16}{*}{CNN}  & \multirow{4}{*}{64} & \multirow{2}{*}{MNIST}&\ Train & 99.91 & 17.31 & 73.29 & 99.99\\
  &  &\ &\ Validation& 98.50 & 10.67 & 10.20& 09.73\\
  &  & \multirow{2}{*}{FashionMNIST}&\ Train & 96.53 & 61.55 & 61.04 & 99.85\\
  &  &\ &\ Validation& 89.58 & 10.33 & 09.72& 09.83\\
  & \multirow{4}{*}{128} & \multirow{2}{*}{MNIST}&\ Train & 99.96 & 18.36 & 99.26 & 100\\
  &  &\ &\ Validation& 98.63 & 10.62 & 10.10& 09.23\\
  &  & \multirow{2}{*}{FashionMNIST}&\ Train & 97.26 & 91.38 & 86.39 & 99.99\\
  &  &\ &\ Validation& 91.63 & 09.82 & 09.53& 09.70\\
  & \multirow{4}{*}{256} & \multirow{2}{*}{MNIST}&\ Train & 99.97 & 33.93 & 100 & 100\\
  &  &\ &\ Validation& 98.62 & 09.98 & 09.82& 09.88\\
  & & \multirow{2}{*}{FashionMNIST}&\ Train & 97.74 & 99.67 & 99.98 & 100\\
  &  &\ &\ Validation& 91.52 & 10.23 & 09.55& 09.68\\
  & \multirow{4}{*}{512} & \multirow{2}{*}{MNIST}&\ Train & 99.99 & 56.22 & 100 & 100\\
  &  &\ &\ Validation& 98.70 & 10.10 & 09.82& 10.22\\
  & & \multirow{2}{*}{FashionMNIST}&\ Train & 98.05 & 99.99 & 99.94 & 100\\
  &  &\ &\ Validation& 90.37 & 09.52 & 10.00& 10.38\\
  \hline

\multirow{8}{*}{denseNet} & \multirow{2}{*}{64} & \multirow{2}{*}{cifar10}&\ Train & 100 & 43.17 & 52.48 & 100\\
  &  &\ &\ Validation& 86.52 & 10.40 & 09.46& 09.90\\
  & \multirow{2}{*}{128} & \multirow{2}{*}{cifar10}&\ Train & 99.84 & 85.58 & 93.17 & 100\\
  &  &\ &\ Validation& 85.32 & 10.02 & 10.36& 10.04\\
  & \multirow{2}{*}{256} & \multirow{2}{*}{cifar10}&\ Train & 99.76 & 99.76 & 100 & 100\\
  &  &\ &\ Validation& 83.72 & 09.66 & 10.40& 10.20\\
  & \multirow{2}{*}{512} & \multirow{2}{*}{cifar10}&\ Train & 99.70 & 99.85 & 100 & 100\\
  &  &\ &\ Validation& 81.82 & 10.14 & 10.12& 10.24\\

\end{tabular}
}
\end{table} 

\subsection{Hypotheses}

Before going into the analysis, we believe it is beneficial to  state the key hypotheses that will be evaluated in the following sections. 
\begin{enumerate}

\item As memorization is likely to involve fewer nodes specializing on rare input features, we assume this will result in a lower overall firing rate.
\item There could be a tendency for the Poisson model to better fit for generalizing and random models, due to potential non-stationary memorizing nodes.
\item We expect random and generalizing models to behave rather similarly (due to the aims of standard weight-initialization methods, see \cite{glorot_understanding_2010}) and memorizing nodes to deviate more strongly.
\item It is not clear how the firing rates within a layer will be distributed, but we might observe a higher variability within the memorizing networks, as some nodes might stay closer to initialized behavior while others move away.
\item For wider models, we expect strong overparameterization (especially for MNIST/FashionMNIST) and therefore more nodes without function. This might lead to a kind of 'dilution effect', meaning that measures on the memorizing and generalizing networks are more difficult to distinguish. 
\item It is unclear what effect the depth of different models have on the observed effect. Memorization tends to happen most strongly in later layers \cite{stephenson_geometry_2021}, which is why we observe only the last layer in our experiment. Depth likely affects the amount of memorization that has to be done by that last layer, and this might affect the results.
\item As we are looking only at the thresholded spike trains, we expect activation patterns to be less dependent on specific input. We therefore hope that the patterns are similar regardless of whether they are generated on the known training or the unknown testing data. 
\item We hope that the measured firing rates are stable over training under one condition, with clearly visible 'change' points when the networks change from generalization to memorization.
\end{enumerate}

\subsection{Experimental results}
We split the presentation of the experimental results in three parts. First, we look at how well the observed data is being modeled as a Poisson Process by discussing the key assumptions of these processes and how well they fit the data. Second, we  investigate how much Poisson Process parameters, like the mean firing fate ($\lambda$)  and different measures of variation, vary between generalizing and memorizing models. In the third part, we observe these measures during the training process to investigate whether they could be used to spot memorization before it occurs. If this is the case, the proposed measures could find practical applications, e.g. as an (additional) early stopping criteria or as a penalty term in regularization.
\subsubsection{Key assumptions \& Fano Factors}
Lacking a theoretical test for how well the Poisson process model describes the observed data, some researchers calculate Fano factor (F) of an observed process as a test statistic \cite{eden_drawing_2010} \cite{rajdl_fano_2020}.  The Fano factor of a counting process $N$ is defined as $F=\frac{Var(N(w))}{E(N(w))}$ over a time window $w$ and can be seen as a measure of variability for the process. An interesting feature of the theoretical Poisson processes is that, as their variance and mean are equal,  their theoretical $F=1$  \cite{cox_renewal_1962}.  Based on that, Fano factors close to one are sometimes taken as indicators of a good theoretical fit to a Poisson process model. \cite{eden_drawing_2010} presents a statistical test based on the observation that the sampling distribution of the Fano factor tends to a gamma distribution. We calculated F for the observed data using a window of size 100 and implemented the methodology described in \cite{eden_drawing_2010}  to  generate significant levels and p-values for each node in each network. Table \ref{table:mf} presents the mean F for each network together with the results of this hypothesis test. Most spike trains have a Fano factor significantly below 1, which suggests that the variance is lower than expected by the model. Memorizing ANNs show a tendency to have higher Fano factors than random or generalizing ANNs and are in many cases significantly close to 1. This could be taken as an indicator that those network’s activations are more suitable to be modeled by a Poisson process. However, known issues of Fano factors are its dependency on parameters like the window size and the value of $\lambda$ \cite{rajdl_fano_2020}. Lower firing rates ($\lambda$) do lead to inflated Fano factors, which makes them difficult to interpret if looking at models with different firing rates. Indeed, in our experiments, we see a near linear relationship between the mean Fano factor and the mean Firing rate of a network (see figure \ref{fig:RelFF}). \\
In the absence of a reliable test for model fit, we argue it to be more important to consider how well the key assumptions underlying the Poisson process fit the observed data.

\begin{figure}[h!]
    \centering
    \subfigure[]{\includegraphics[width=0.49\textwidth]{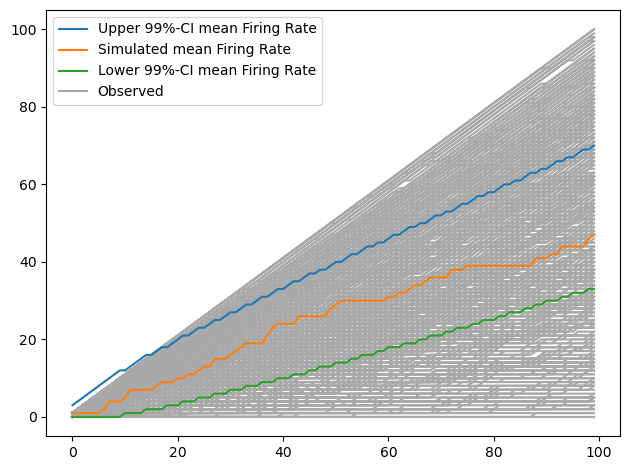}} 
    \hfill
    \subfigure[]{\includegraphics[width=0.49\textwidth]{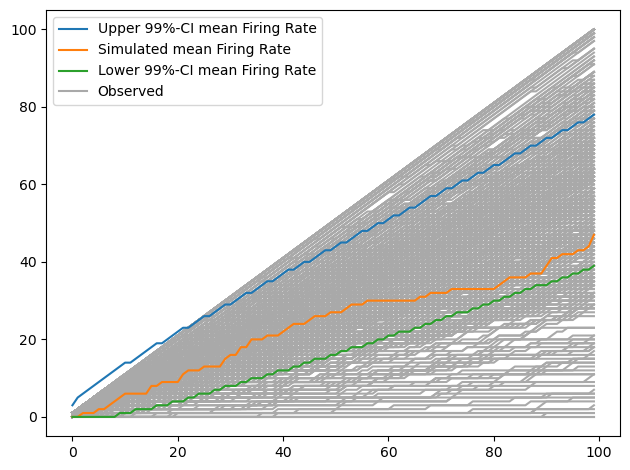}} 
    \hfill
    \subfigure[]{\includegraphics[width=0.49\textwidth]{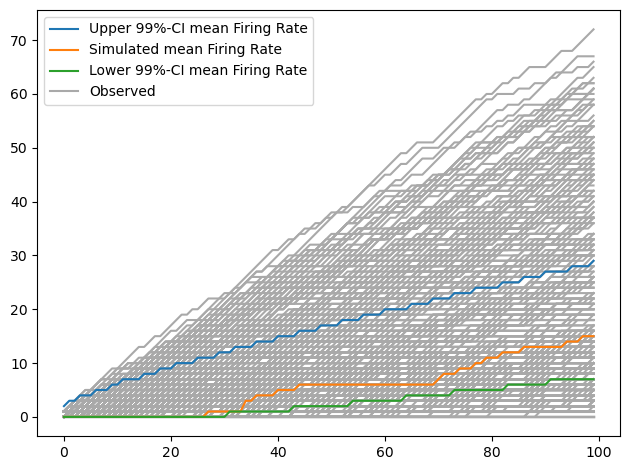}} 
    \subfigure[]{\includegraphics[width=0.49\textwidth]{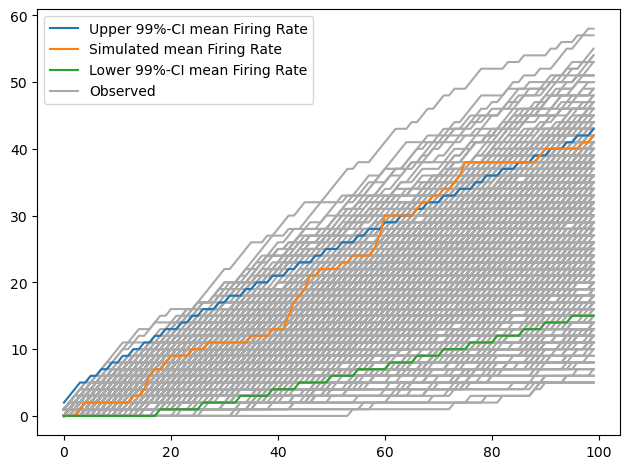}} 

\caption{Observed cumulative Spiketrains and simulated Poisson Process set to the Mean Firing Rate of each ANN. The examples show results from the 512-nodes MLP on the testing data from MNIST. Each gray line shows the cumulated spike train generated by one node in the ANN over 300 samples.  a) shows the randomly initialized ANN b) the ANN after training for generalizing, c) the same network after retraining on the shuffled dataset and d) a randomly initialized ANN trained for memorization} \label{fig:Poi}
\end{figure}

The key assumptions underlying the Poisson Process model are that spikes occur (1) one-at-a-time and are  (2) independent as well as (3) stationary. Above, we proposed that the spiking patterns of ANN generated by the described procedure would fit these assumptions well and better than real-life neural spikes.\\
The first assumption naturally fits our generated data, because any node in the ANN either activates or not. As we model one Poisson process per nodal spike train, spikes occur one-at-a-time. The activation of any node of a non-recurrent network is in principle also mechanistically independent of previous activation. However, the sampling process to generate input data might induce statistical dependency. If e.g. the inputs are ordered, nodes corresponding to specific classes might show rhythmic spiking activity. We therefore randomized the ordering of samples to decrease such induced dependencies. There is to the best of our knowledge no formal statistical test for independence in spike series, but the lagged autocorrelation can be an indicator \cite{kramer2016case}. We therefore conducted independent Ljung–Box tests \cite{ljung_measure_1978} on the spike trains generated by each node and combined the calculated p-values using the Pearson-method \cite{heard_choosing_2018} as implemented in the statsmodels package \cite{seabold_statsmodels_2010}. We didn't observe large, significant autocorrelations in any of the spike trains.
The assumption of stationarity is another case which is given mechanistically (the spike producing ANN doesn't change after training), but which could be violated by the interaction between model and input data. For example, for memorizing ANNs, a node could in principle  just respond to a singular, memorized sample in the training data. In this case, its activation probability would go to 0, after activating once. As with independence, stationarity is difficult to reliably test against. Algorithms such as \cite{messer_multiple_2014} for change point detection do give some measure for the degree of non-stationarity, but its results are difficult to interpret and do not equal a statistical hypothesis test. We did a test against unit-root (one sign on non-stationarity) using the Augmented Dickey-Fuller test as implemented in \cite{seabold_statsmodels_2010} and didn't find any signs for non-stationarity under any condition. We do consider the stationarity of observed activations as an assumed premise. \\
Seeing that these three key characteristics describe well the activation behavior of ANN, we argue that the Poisson process should be a suitable baseline model of observed activation patterns. Figure \ref{fig:Poi} shows visual examples of observed spike trains and simulated Poisson processes. Judging by visual inspection, the behavior of the observed nodes seems to be not too far off from the simulated Poisson process, although the variability of individual spike trains might be lower than the theoretical Poisson process. Furthermore, the variability between nodes is higher than the confidence intervals for the simulated mean process. For simulation purposes, one process-parameter would not be enough to simulate the observed ANN.

\subsubsection{Poisson Process Parameters indicating Memorization}
\begin{figure}[h!]
    \centering
    \subfigure[]{\includegraphics[width=0.49\textwidth]{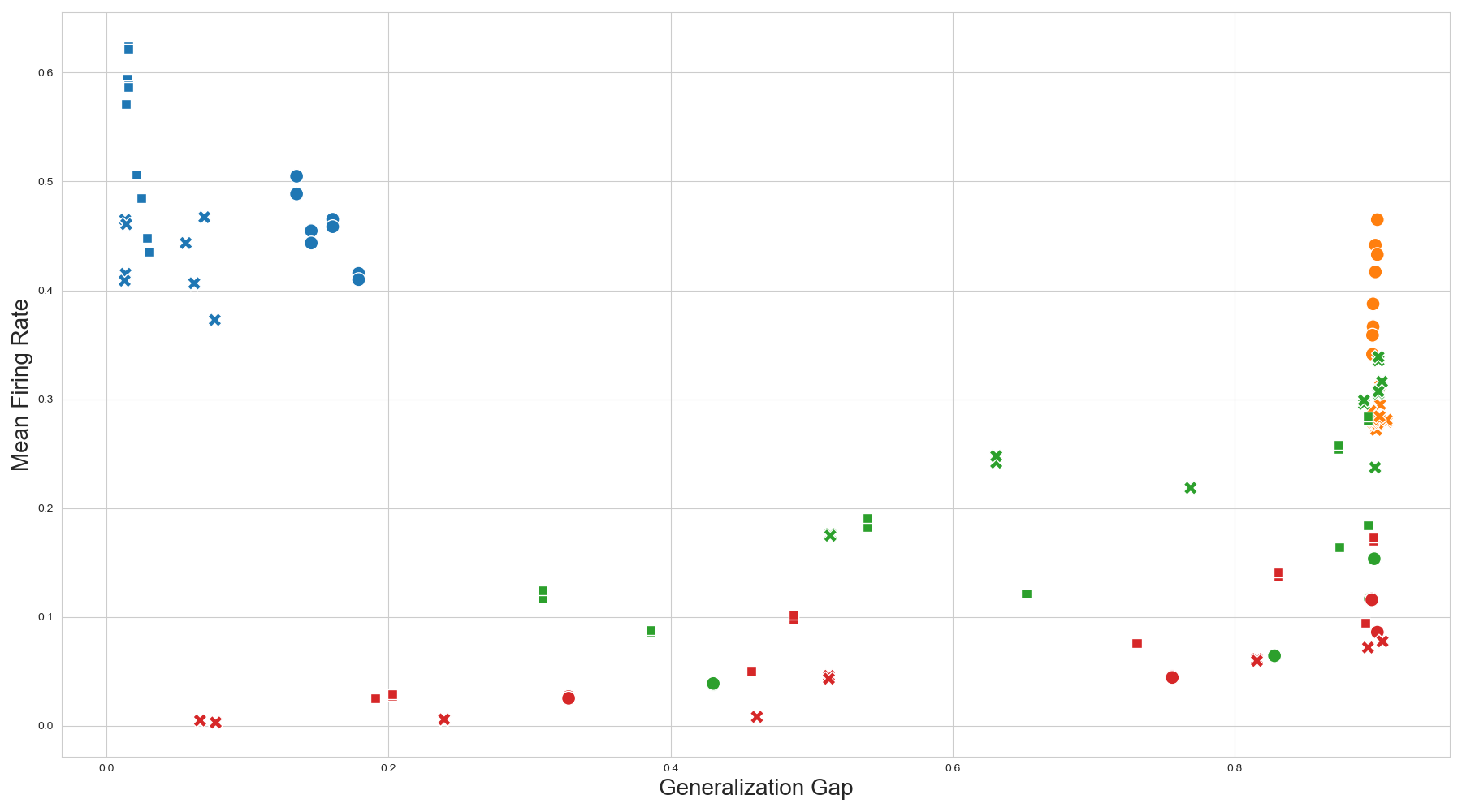}} 
    \hfill
    \subfigure[]{\includegraphics[width=0.49\textwidth]{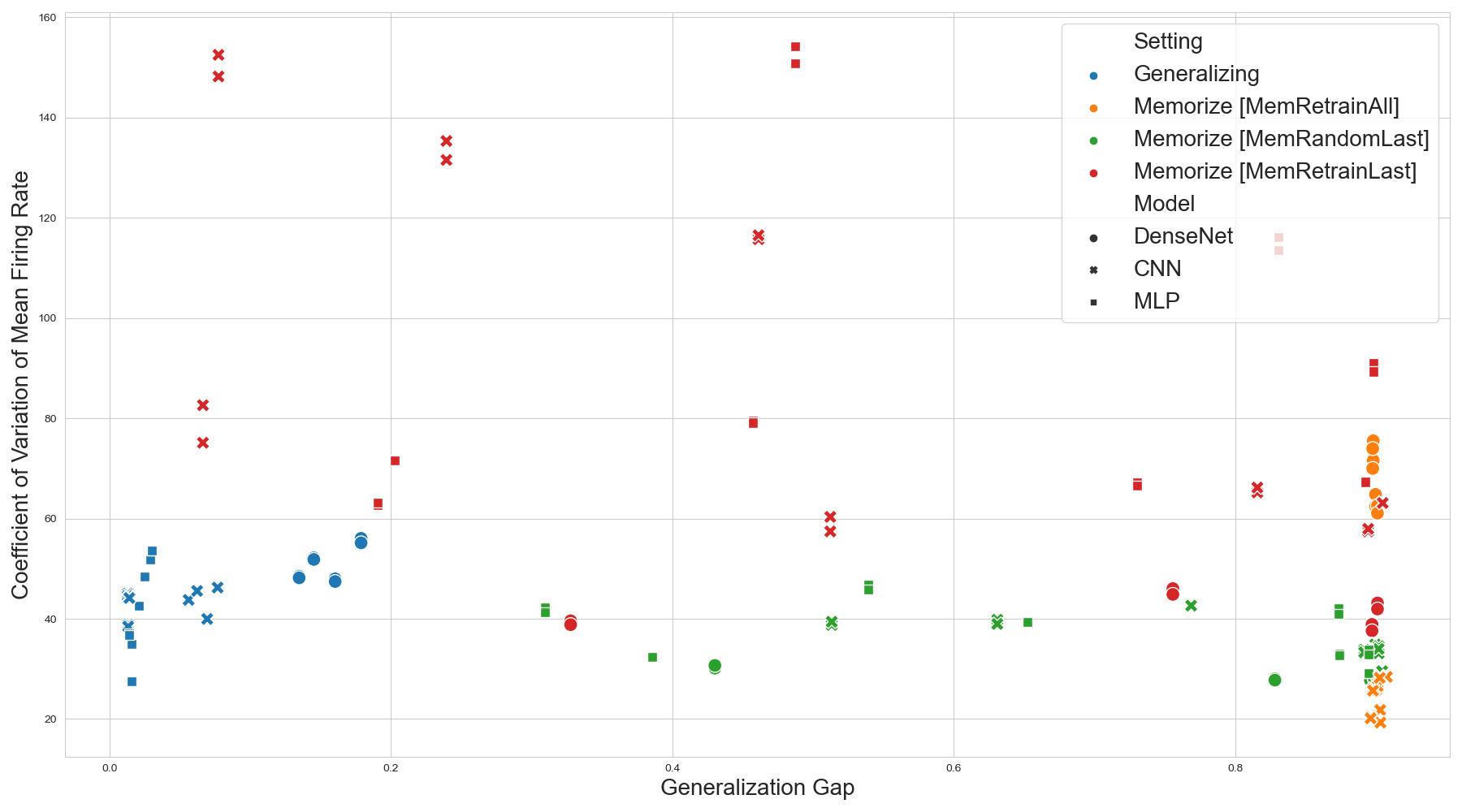}} 
    \hfill
    \subfigure[]{\includegraphics[width=0.49\textwidth]{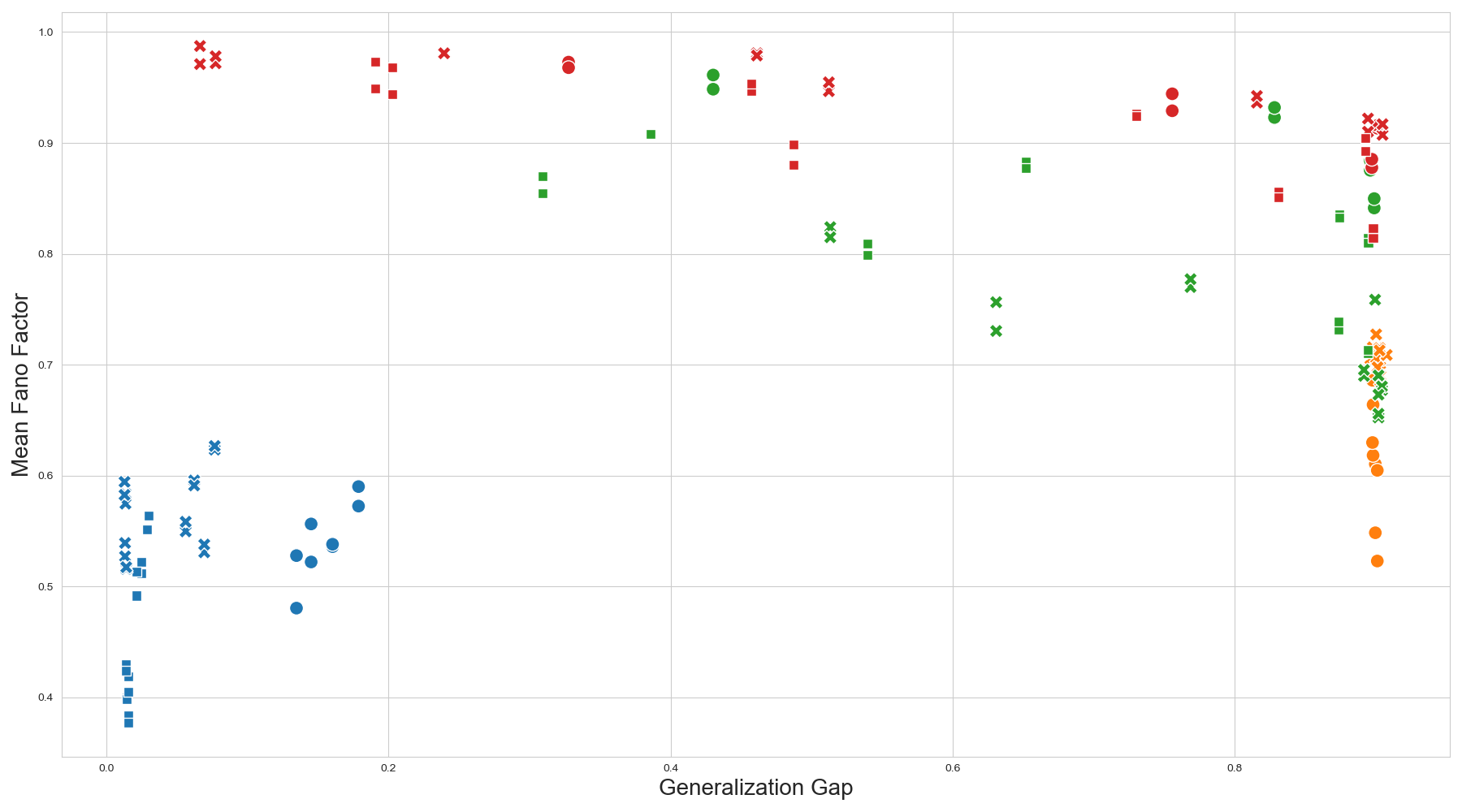}} 
    \hfill
    \subfigure[]{\includegraphics[width=0.49\textwidth]{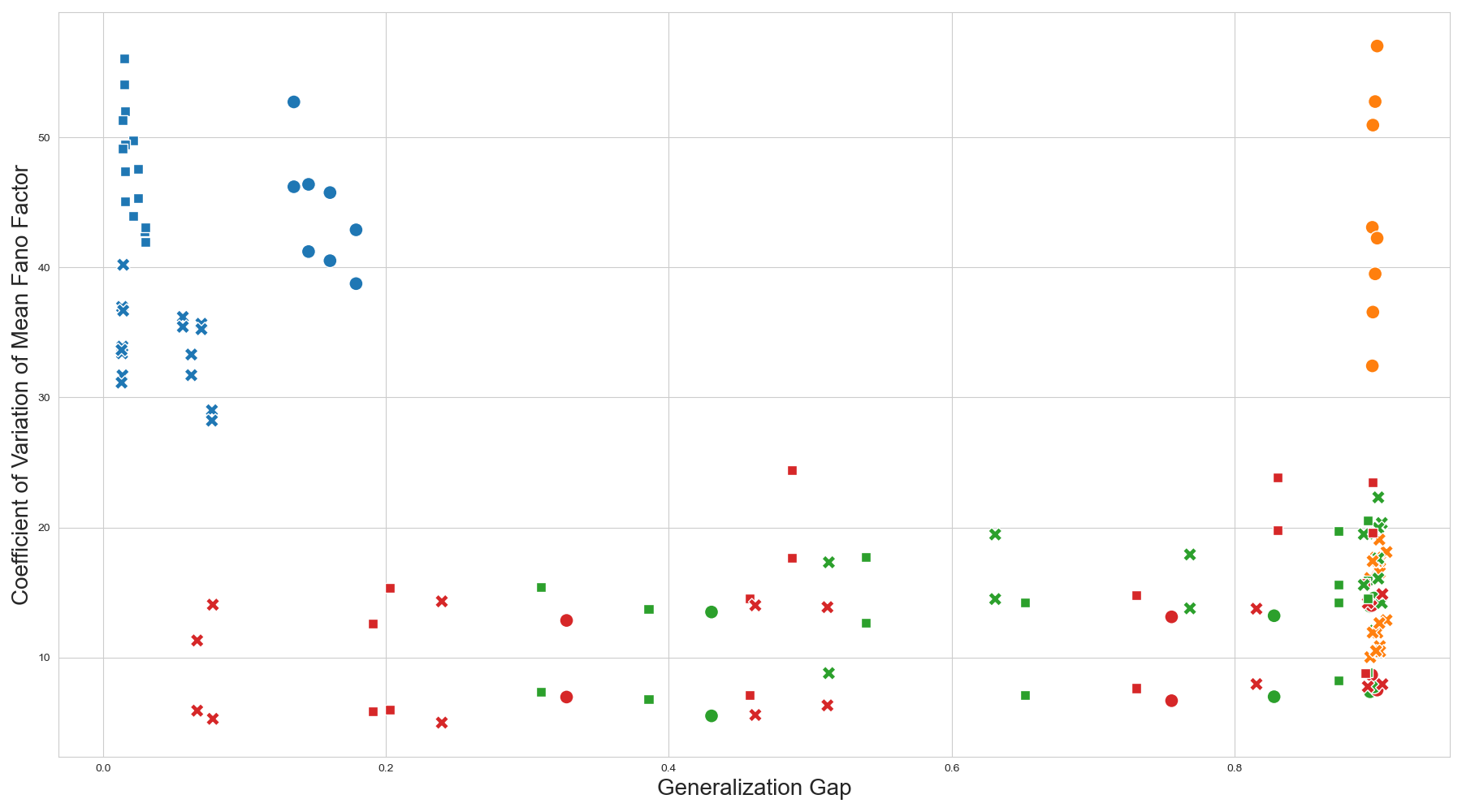}} 

\caption{Relationship between key statistics and generalization gap of the same model. On the x-axis the generalization gap (difference in accuracy between training and validation set), on the y-axis the respective statistic. Different colors represent different training settings, different shapes in the different datasets. (a) The mean of the firing rate of each node in the observed layer. (b) The coefficient of variation of the firing rate. (c) The mean of the Fano factor of each node in the observed layer, calculated using a window of 100 samples. (d) The coefficient of variation of the Fano factor. } \label{fig:GenGap}
\end{figure}

\begin{figure}[h!]
    \centering
    \subfigure[]{\includegraphics[width=0.49\textwidth]{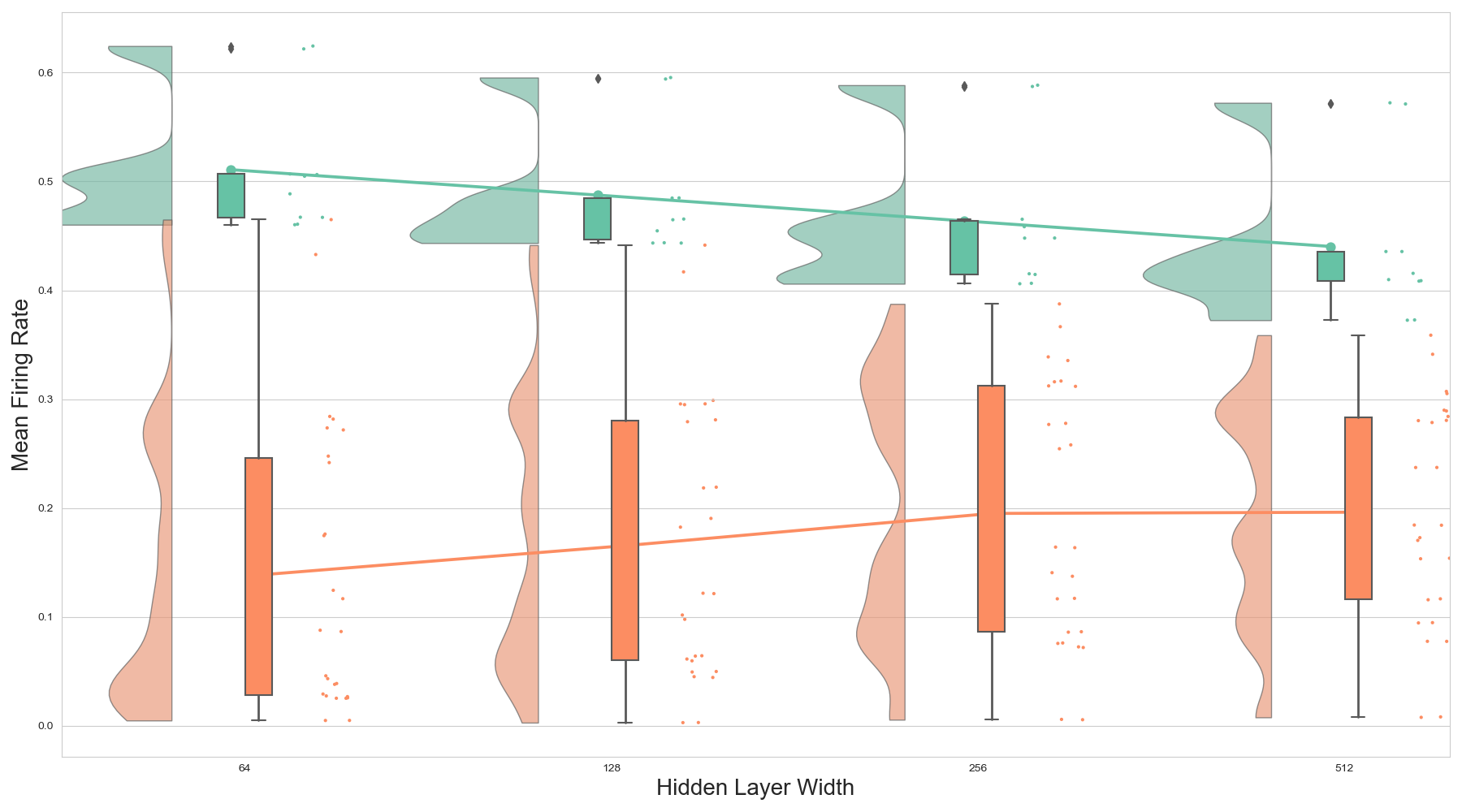}} 
    \hfill
    \subfigure[]{\includegraphics[width=0.49\textwidth]{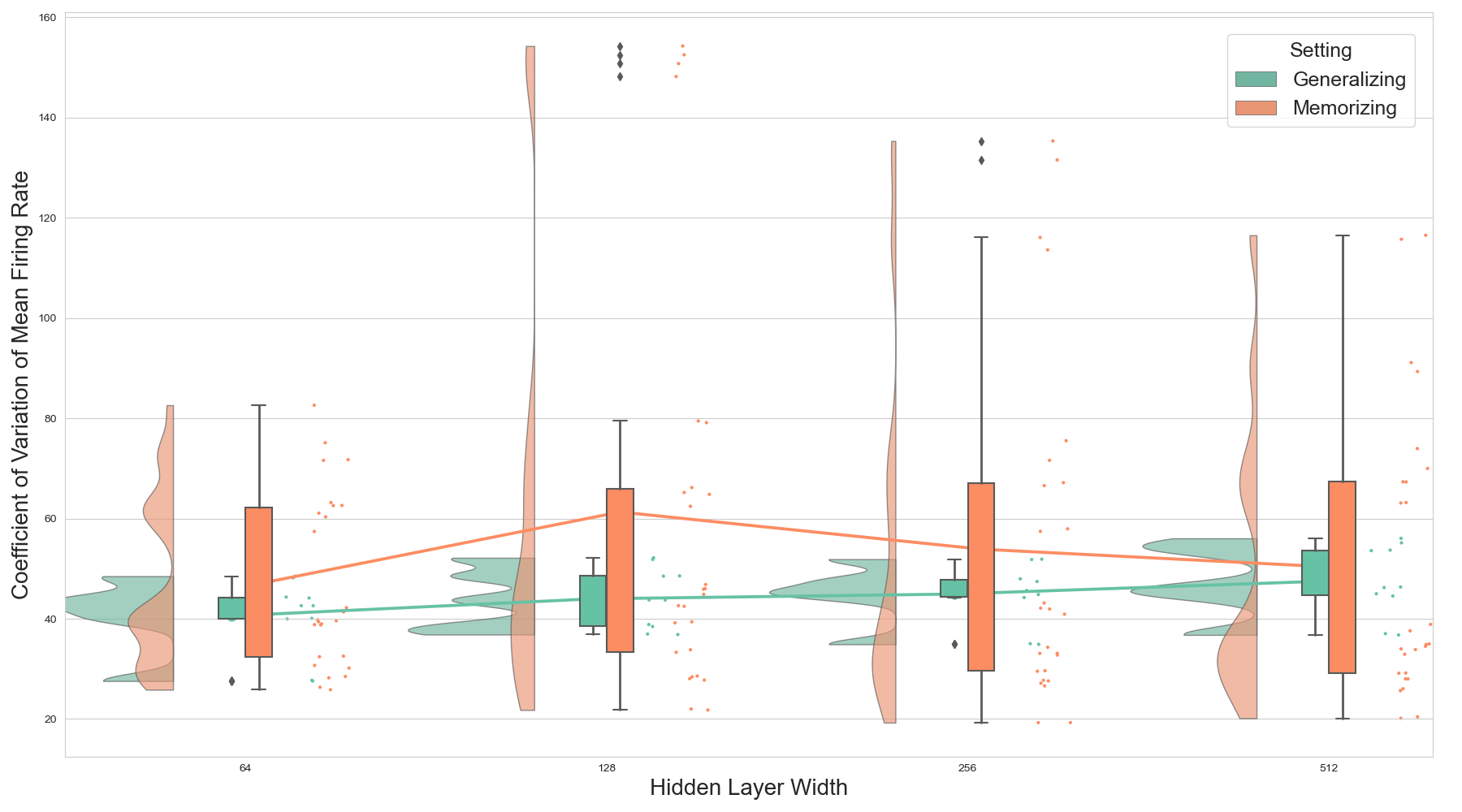}} 
    \hfill
    \subfigure[]{\includegraphics[width=0.49\textwidth]{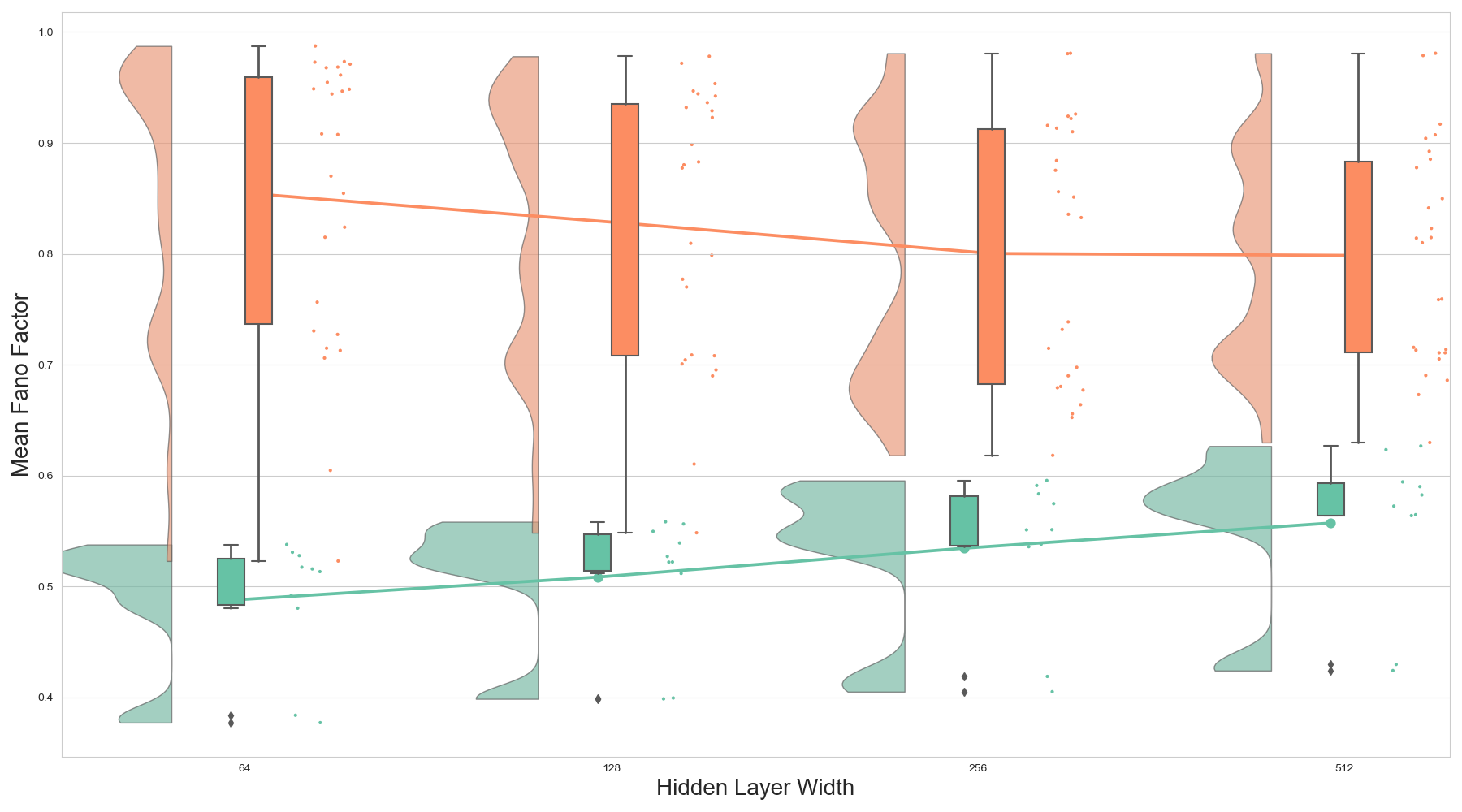}} 
    \hfill
    \subfigure[]{\includegraphics[width=0.49\textwidth]{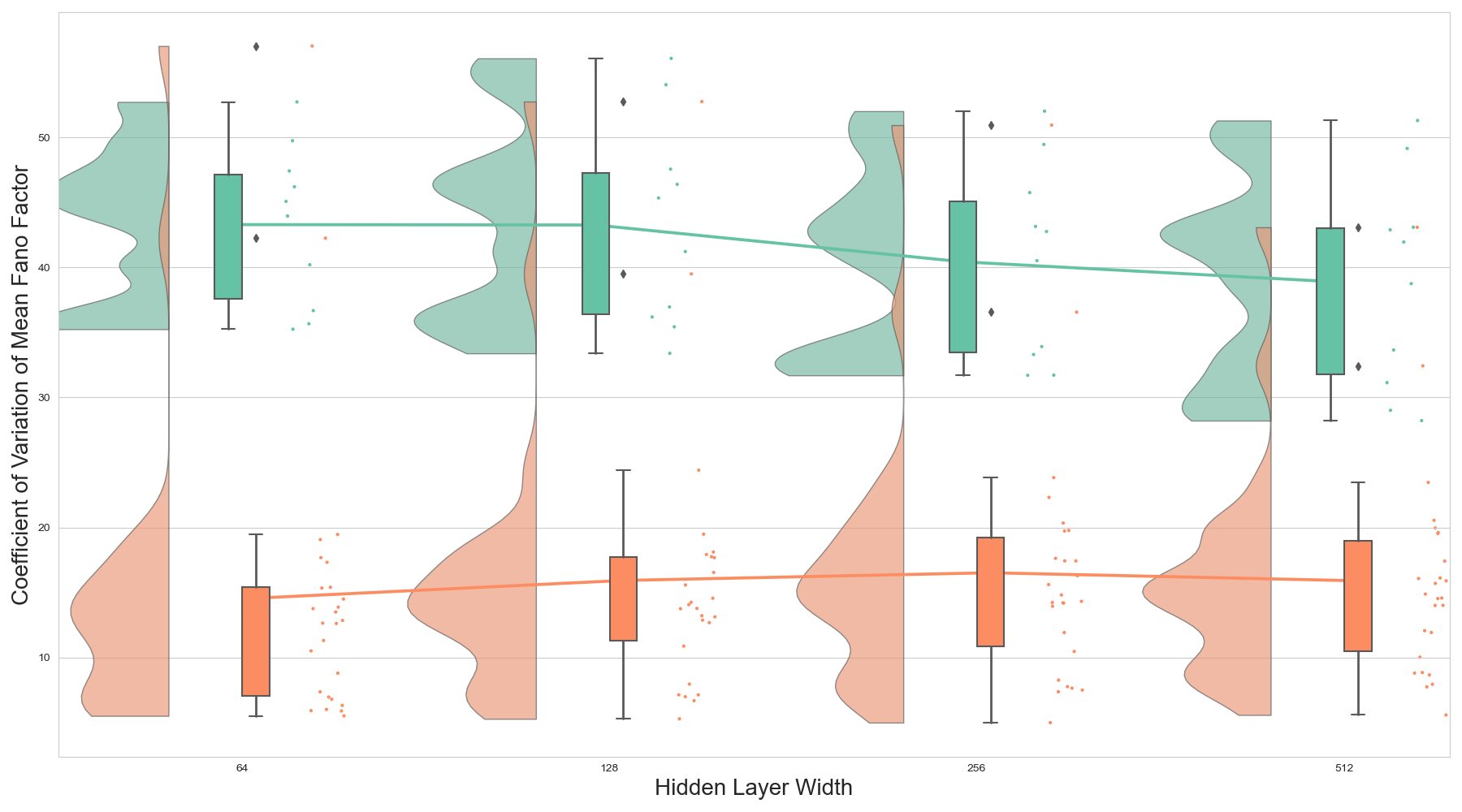}} 

\caption{Rainplots showing the differences between key statistics when Generalizing \& Memorizing plotted against width of the observed layer. (a) The mean of the firing rate of each node in the observed layer. (b) The coefficient of variation of the firing rate. (c) The mean of the Fano factor of each node in the observed layer, calculated using a window of 100 samples. (d) The coefficient of variation of the Fano factor.} \label{fig:Diff}
\end{figure}

\begin{figure}[h!]
\centering
    \includegraphics[width=0.85\textwidth]{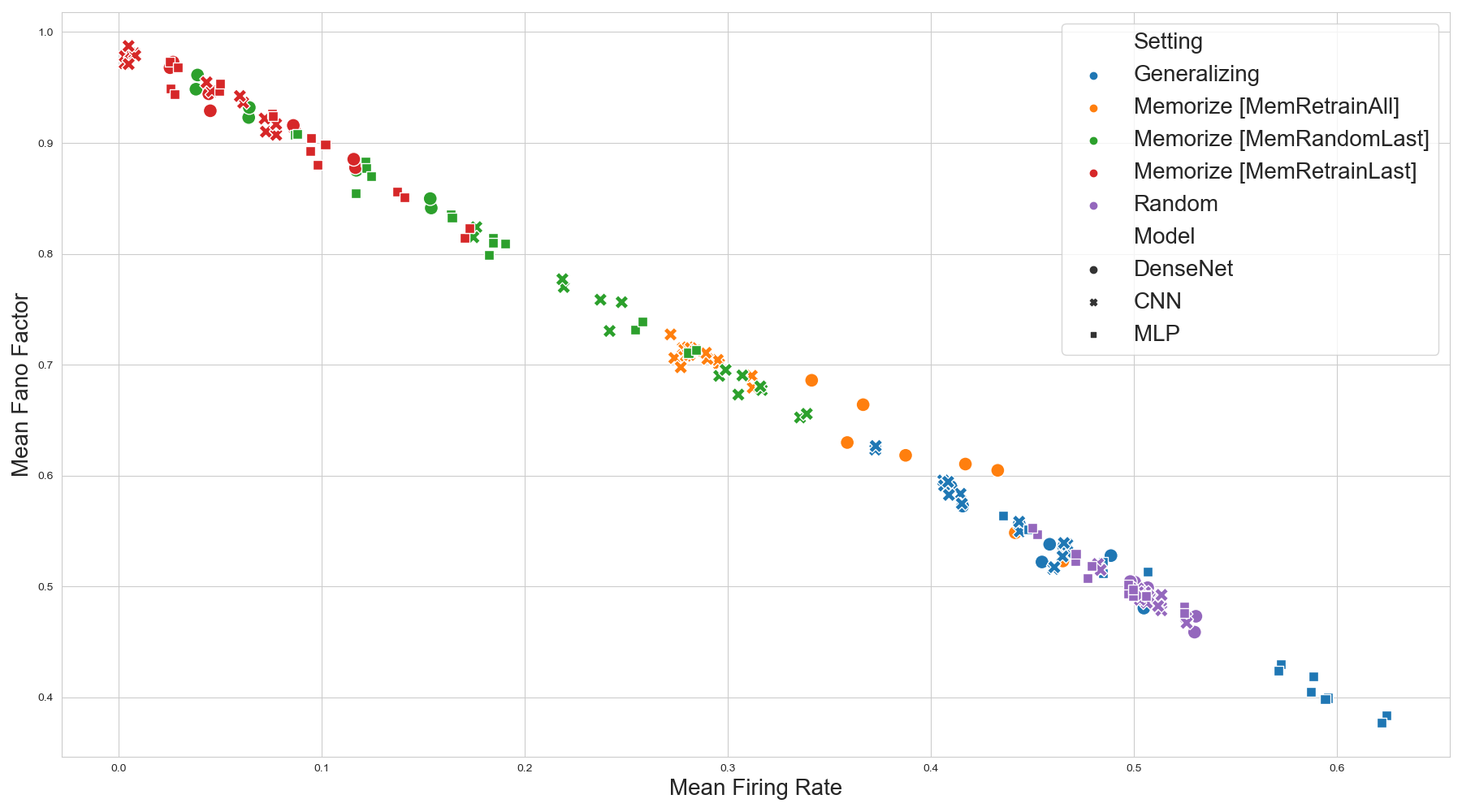}
\caption{Relationship between Mean Fano Factor and Mean Firing Rates} \label{fig:RelFF}
\end{figure}

We are going to discuss the  general distribution of the observed $\lambda$ for each ANN as well as 4 key parameters calculated from the fitted Poisson process models for each ANN and investigate how they differ between memorizing and generalizing networks. As mentioned above, we assume systematic differences to exist and these parameters could be an important metric for model training, if they can be useful for discriminating memorization from generalization.\\
\\
Initially, we compared the overall distribution of the firing rate (FR) between the training and test data and the different settings (Generalization, Memorization, Random). For this, we used the Kolmogorov-Smirnov test (KS-test) for equality of continuous probability distributions. We interpret the results in light of the leading hypotheses in the following; Table \ref{FR-Table} shows a summary of results.\\
As initially assumed (hypothesis 3), the generalizing and random ANNs have a tendency to have similar FR distributions. They do show close MFR values in all settings and in many cases significantly similar distributions (see table \ref{FR-Table}, rows marked with $\heartsuit$ show models with significant similarities between generalizing and random state). Only some  wider models with more nodes observed do start to show stronger deviations from the random state (hypothesis 5).
In contrast to that, the memorizing model’s FR distribution differs significantly from both random and generalizing models in nearly all cases. Only the larger DenseNet models, and only under the condition that all layers are used for memorization [MemRetrainAll], show significant similarity to Random or Generalizing ANN (see table \ref{FR-Table}, rows marked with $\clubsuit$ show models with significant similarities between memorizing and generalizing state, rows marked with $\diamondsuit$ between memorizing and random state; this is only the case for deeper networks when all layers are used for retraining).\\ 
Over all the tested models, we couldn't find any significant differences between the distribution of spike trains generated on the known training data or the unknown test data. This similarity is very useful, as it indicates that a separated test set would not be strictly necessary to evaluate MFR-based indicators for memorization in an ANN.\\
We also observe some indication that the width of the observed layer affects the MFR. The mean firing rate distribution of generalizing and memorizing networks show some tendency to move closer together when observing wider layers. MFR gets consistently smaller for generalizing networks and sometimes larger for the memorizing networks. This could be due to a dilution effect postulated in hypothesis 5. 
A potential depth effect as postulated in hypothesis 6 might be visible in the MemRetrainAll condition. We see that FR distribution differs less between memorizing and generalizing networks in the case of the deeper networks, as long as all layers are used for memorization. When layers before the observed layer are frozen, the FR distribution stays distinct. \\
\\
To test the applicability of Poisson process parameters as single variable diagnostic indicators of memorization in a network, we evaluated for simple indicator variables. For each node of each ANN, we first calculate the firing rate (FR), which is simply the $\lambda$ parameter of the Poisson process, and the Fano factor (F), as described in the sections above. To gain simple indicators for each network, we then summarize these as the Mean Firing Rate (MFR) and Mean Fano factor (MF). To get insight into the distributional characteristics of these indicators, we also calculate their respective Coefficient of Variation (e.g. $CV = \frac{\sigma_{FR}}{MFR}$) of both FR and F over the observed nodes of a network.\\
Figure \ref{fig:Diff} shows the difference between those indicators calculated on the same network under generalizing conditions and memorizing conditions. Figure \ref{fig:GenGap} shows how those indicators relate to the Generalization Gap (measured as $acc_{train} - acc_{validation}$). \\
We observe that MFR (\ref{fig:GenGap} (a)) and MF (\ref{fig:GenGap} (c)) show both distinct differences between the generalizing and the memorizing settings. As MFR and MF are (in our experiments, see figure \ref{fig:RelFF}) linearly related, a similar pattern was expected. Overall, memorizing networks appear to show lower firing rates and higher Fano factors than the generalizing networks. Only networks, where memorization might take place in layers before the observed on, the MemRetrainAll condition, show some overlap in MFR/MF-values with the generalizing networks. This could again be caused by memorization happening in the previous layers, which weren't observed in our experiment (hypothesis 6).\\
The Variation (measured as CV) for both MFR and MF show stronger deviating patterns. The Coefficient of Variation for MFR, Figure \ref{fig:GenGap} (b), does show some clustering of the different models and memorization methods, but the CV itself seems to be not a strongly distinguishing factor between memorizing and generalizing models. This goes against hypothesis 4.
The CV for MF, Figure \ref{fig:GenGap} (d),  shows more stable and therefore interesting behavior. We observe similar values, regardless of the generalization gap (= how much the network memorized), with MemRetrainAll being the only outlier. This makes CV MF a promising candidate as a differentiating indicator between memorization and generalization in training. However, the behavior of CV MF is difficult to explain, and it could also be an artifact of the estimation procedure.\\
The method of memorization seems to affect the relationship between generalization gap and MFR/MF, while differences in ANN models seem to have no impact. This shows in slightly different slopes for the different settings. The slope is noticeable so far, that it is moving memorizing ANNs with a higher generalization gap (=higher number of memorized samples) to slightly higher/lower values of MFR/MF. The values get therefore slightly closer to the value of generalizing models, without reaching the same level for most models. While this potentially contradicts the expectation and observation that MFR/MF differs with higher memorization, this could also be due to the fact that wider models more easily memorize and the postulated dilution due to width. Figure \ref{fig:Diff} also examines the effect of layer width on the investigated parameters. It is important to understand, whether the estimated parameters differ stable between memorization based learning and generalization based learning or whether further memorization leads to a higher similarity in these parameters. For this reason, we evaluate how the parameters change during training in the next section. 

Figure \ref{fig:Diff} shows the distribution of the four parameters investigated over all observed models and settings and shows how it is affected by the width of the observed layer. For MFR and MF, we see a rather strong effect of layer width. As assumed in hypothesis 3, the Poisson process parameter for generalizing and memorizing ANNs move closer together, if more nodes are observed. As mentioned above, the next section will discuss if this is caused by the better performance of the wider models, or if it is more likely caused by the hypothesized width effect. \\
Each parameter in the generalizing setting shows a multi-modal distribution with three clearly visible peaks, indicating that the three different models have a strong effect on the parameter. Under memorization, we do observe multi-modality, but do not see such strong peaks, indicating, again, that the difference between specific models is not as pronounced.\\
Regarding the overall direction, we see again that MFR and CV MF have relatively stable lower values under memorization, MF higher values and CV MFR itself does not provide a clear picture, although its distribution deviates strongly between both conditions.

\subsubsection{Poisson Process Parameter during training}
\begin{figure}[h!]
    \centering
    \subfigure[]{\includegraphics[width=0.49\textwidth]{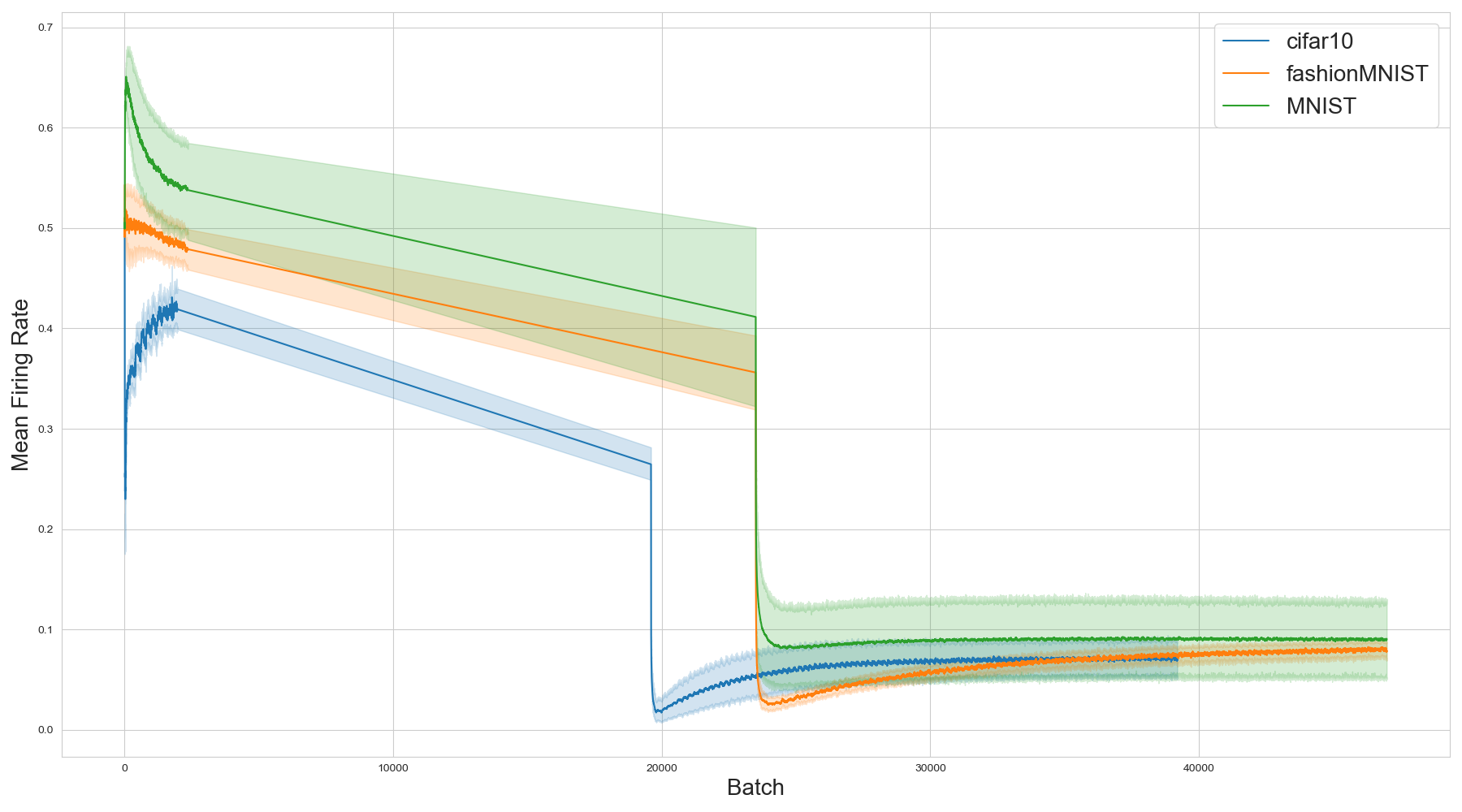}} 
    \hfill
    \subfigure[]{\includegraphics[width=0.49\textwidth]{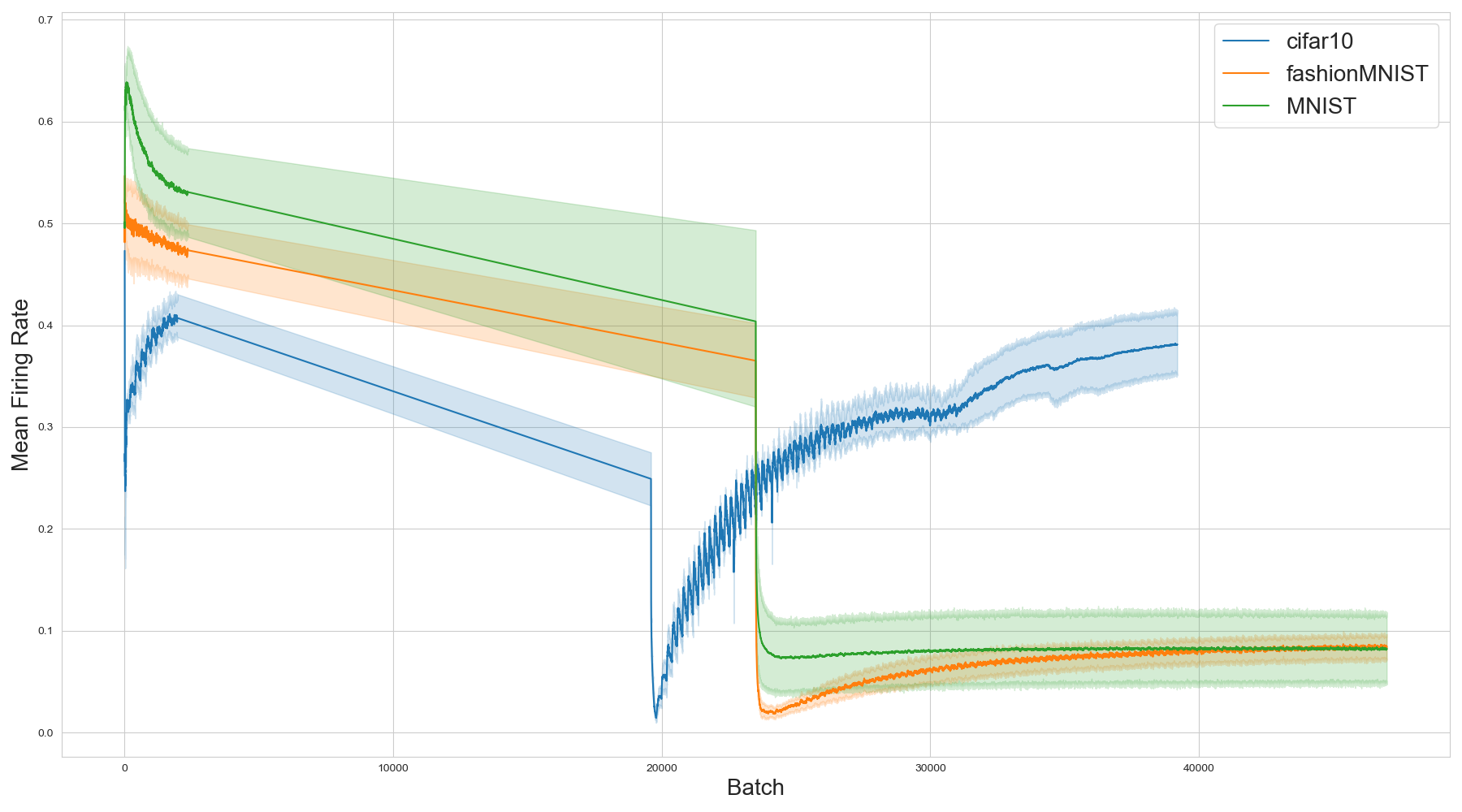}}

\caption{Mean Firing Rate during Training. On the x-axis, the different batch number, on the y-axis, Mean Firing Rate (MFR). Results are grouped by dataset, shown by different colors. Within each group, we show the mean MFR together with confidence intervals. (a) Retraining setting with all layers but the observed frozen [MemRetrainLast]. (b) Retraining setting where all layers are allowed to memorize [MemRetrainAll].} \label{fig:Dur}
\end{figure}

As established in the previous section, we found some indication that descriptive parameters like the mean firing rate and the mean Fano factor do show distinctive differences between memorizing and generalizing networks. Some limitations however have been discussed, so far we are not able to rule out that these parameters will begin moving closer together with further memorization. We therefore would be also interested in understanding how these variables change during training. If these measures do remain relatively stable during training, they might be useful indicator variables with practical applications for network pruning, early stopping, transfer learning etc. \\
We trained the same networks as above for 10 epochs on data with the correct labels (generalization) and afterward continue training the network on the dataset with randomized labels (memorization) for 100 epochs. During the training process, we let the network generate short spike trains on a small sample of the training and test data (100 samples) at the end of each batch.\\
Figure \ref{fig:Dur} shows the development of the Mean Firing Rate during training of all networks. Starting from the randomly initialized ANNs with a mean firing rate close to 0.5, at the beginning of training, the MFR first drops strongly, then slowly rises with some variation again. After the first epoch, we observe a mostly stable MFR, with a slight downward slope. The form of the rise and slope difference between each dataset, but does not show too much variation between variation networks and sizes.\\
After exchanging the original training dataset for the shuffled version, we again notice a quick drop in the MFR, afterward a slow increase, followed by a stable value of the MFR. The only outlier is the already known case of MemRetrainAll, Figure \ref{fig:Dur} (b), where we notice an overall less stable behavior on cifar-10 with a continuous rise of the MFR. As already noted earlier, this is likely due to most of the memorization happening in unobserved, earlier layers.\\
We observed stable values of MFR during continuous memorization. These observations indicate that the observed slopes in Figures \ref{fig:GenGap}(a,c) are indeed caused by the width effect. The increase of MFRs by increasing generalization gap are less likely to be caused by a higher degree of memorization, as MFR stays stable during continuing memorization in this experiment. The pronounced behavior of the MFR during initial batches when observing new data is also an interesting phenomenon, with potential practical applications.

\begin{table}[htp]

    \centering

    \caption{\label{FR-Table} MFR: Mean Firing Rate; CV: Coefficient of Variation; KS-Test: results of Kolmogorov-Smirnov test for goodness of fit, Significant (p-value > 0.00027 [ $\alpha$-level 0.05 with Bonferroni Correction for 128 independent comparisons]) similarities are marked with symbols; $\clubsuit$ : between Test/Train, $\diamondsuit$ :  between Memorization/Random, $\heartsuit$ :  between Generalization/Random, $\spadesuit$:  between Generalization/Memorization, *only for condition MemRetrainAll }
\resizebox{\textwidth}{!}{
\begin{tabular}{c|c|c|c|c|c|c|c|c|c|c|c|c|c|c}
\hline
\multirow{2}{*}{Model} & \multirow{2}{*}{Width} & \multirow{2}{*}{Data} & \multirow{2}{*}{Data split} & \multicolumn{2}{c}{Generalizing} & \multicolumn{2}{c}{MemRetrainLast} & \multicolumn{2}{c}{MemRandomLast} & \multicolumn{2}{c}{MemRetrainAll} & \multicolumn{2}{c}{Random} & \multirow{2}{*}{KS-Test} \\
&&&&  FR & CV   & FR& CV & FR&CV & FR&CV & FR&CV\\
\hline

\multirow{16}{*}{Dense} & \multirow{4}{*}{64} & \multirow{2}{*}{MNIST}&\ Train & 0.62 & 27.80 & 0.03 & 72.16 & 0.12 & 41.70 & - & - & 0.45 & 60.92 & \multirow{2}{*}{$\clubsuit\heartsuit$} \\
 &  &\ &\ Test&  0.62 & 27.93 & 0.03 & 72.30 & 0.12 & 42.54 & - & - & 0.45 & 61.64 &  \\
 &  & \multirow{2}{*}{FashionMNIST}&\ Train & 0.51 & 42.94 & 0.03 & 63.67 & 0.09 & 32.67 & - & - & 0.52 & 61.74 & \multirow{2}{*}{$\clubsuit\heartsuit$} \\
 &  &\ &\ Test&  0.51 & 42.94 & 0.03 & 63.10 & 0.09 & 32.80 & - & - & 0.52 & 61.84 &  \\
& \multirow{4}{*}{128} & \multirow{2}{*}{MNIST}&\ Train & 0.59 & 37.00 & 0.10 & 151.33 & 0.19 & 46.07 & - & - & 0.48 & 55.92 & \multirow{2}{*}{$\clubsuit\heartsuit$} \\
 &  &\ &\ Test&  0.60 & 37.10 & 0.10 & 154.85 & 0.18 & 47.01 & - & - & 0.48 & 56.63 &  \\
 &  & \multirow{2}{*}{FashionMNIST}&\ Train & 0.59 & 37.00 & 0.05 & 79.43 & 0.12 & 39.55 & - & - & 0.47 & 63.23 & \multirow{2}{*}{$\clubsuit\heartsuit$} \\
 &  &\ &\ Test&  0.48 & 48.75 & 0.05 & 79.76 & 0.12 & 39.35 & - & - & 0.47 & 63.26 &  \\
& \multirow{4}{*}{256} & \multirow{2}{*}{MNIST}&\ Train & 0.59 & 34.97 & 0.14 & 113.81 & 0.26 & 40.99 & - & - & 0.50 & 52.65 & \multirow{2}{*}{$\clubsuit$} \\
 &  &\ &\ Test&  0.45 & 51.99 & 0.14 & 116.28 & 0.25 & 42.20 & - & - & 0.50 & 53.27 &  \\
&  & \multirow{2}{*}{FashionMNIST}&\ Train & 0.45 & 51.92 & 0.09 & 235.44 & 0.16 & 32.83 & - & - & 0.48 & 65.84 & \multirow{2}{*}{$\clubsuit\heartsuit$} \\
 &  &\ &\ Test&  0.45 & 51.99 & 0.08 & 67.28 & 0.16 & 33.15 & - & - & 0.48 & 65.97 &  \\
& \multirow{4}{*}{512} & \multirow{2}{*}{MNIST}&\ Train & 0.57 & 36.81 & 0.17 & 89.41 & 0.28 & 32.96 & - & - & 0.51 & 51.47 & \multirow{2}{*}{$\clubsuit$} \\
 &  &\ &\ Test&  0.57 & 37.07 & 0.17 & 91.19 & 0.28 & 33.89 & - & - & 0.51 & 52.14 &  \\
&  & \multirow{2}{*}{FashionMNIST}&\ Train & 0.57 & 36.81 & 0.17 & 89.41 & 0.28 & 32.96 & - & - & 0.51 & 51.47 & \multirow{2}{*}{$\clubsuit$} \\
 &  &\ &\ Test&  0.44 & 53.74 & 0.09 & 67.39 & 0.18 & 29.21 & - & - & 0.50 & 65.20 &  \\

 \hline

\multirow{16}{*}{CNN}& \multirow{4}{*}{64} & \multirow{2}{*}{MNIST}&\ Train & 0.49 & 50.76 & 0.06 & 511.62 & 0.25 & 39.28 & 0.28 & 28.44 & 0.48 & 67.42 & \multirow{2}{*}{$\clubsuit\heartsuit$} \\
 &  &\ &\ Test&  0.46 & 44.68 & 0.00 & 75.70 & 0.24 & 40.12 & 0.28 & 28.71 & 0.48 & 68.27 &  \\
& & \multirow{2}{*}{FashionMNIST}&\ Train & 0.47 & 40.28 & 0.07 & 294.91 & 0.17 & 39.72 & 0.27 & 26.55 & 0.53 & 69.71 & \multirow{2}{*}{$\clubsuit\heartsuit$} \\
 &  &\ &\ Test&  0.47 & 40.39 & 0.05 & 57.87 & 0.18 & 39.09 & 0.27 & 26.04 & 0.53 & 69.87 &  \\
& \multirow{4}{*}{128} & \multirow{2}{*}{MNIST}&\ Train & 0.46 & 38.61 & 0.06 & 527.79 & 0.30 & 33.44 & 0.28 & 28.53 & 0.51 & 61.68 & \multirow{2}{*}{$\clubsuit\heartsuit$} \\
 &  &\ &\ Test&  0.47 & 39.00 & 0.00 & 148.75 & 0.30 & 33.95 & 0.28 & 28.68 & 0.51 & 61.81 &  \\
&  & \multirow{2}{*}{FashionMNIST}&\ Train & 0.44 & 43.94 & 0.09 & 237.65 & 0.22 & 42.79 & 0.29 & 21.89 & 0.50 & 71.63 & \multirow{2}{*}{$\clubsuit\heartsuit$} \\
 &  &\ &\ Test&  0.44 & 43.92 & 0.06 & 65.45 & 0.22 & 42.65 & 0.30 & 22.10 & 0.50 & 71.60 &  \\
& \multirow{4}{*}{256} & \multirow{2}{*}{MNIST}&\ Train & 0.42 & 44.32 & 0.04 & 649.43 & 0.34 & 33.18 & 0.28 & 26.66 & 0.51 & 61.80 & \multirow{2}{*}{$\clubsuit$} \\
 &  &\ &\ Test&  0.41 & 44.91 & 0.01 & 131.77 & 0.34 & 34.39 & 0.28 & 27.19 & 0.51 & 62.28 &  \\
&  & \multirow{2}{*}{FashionMNIST}&\ Train & 0.41 & 45.64 & 0.11 & 475.05 & 0.32 & 29.60 & 0.31 & 19.33 & 0.52 & 68.58 & \multirow{2}{*}{$\clubsuit$} \\
 &  &\ &\ Test&  0.41 & 45.71 & 0.07 & 57.55 & 0.32 & 29.70 & 0.31 & 19.32 & 0.51 & 68.39 &  \\
& \multirow{4}{*}{512} & \multirow{2}{*}{MNIST}&\ Train & 0.41 & 44.59 & 0.06 & 786.89 & 0.31 & 34.02 & 0.28 & 25.69 & 0.50 & 60.78 & \multirow{2}{*}{$\clubsuit$} \\
 &  &\ &\ Test&  0.41 & 45.03 & 0.01 & 115.81 & 0.31 & 34.54 & 0.28 & 26.07 & 0.50 & 61.24 &  \\
& & \multirow{2}{*}{FashionMNIST}&\ Train & 0.37 & 46.27 & 0.10 & 334.29 & 0.24 & 34.91 & 0.29 & 20.17 & 0.51 & 71.02 & \multirow{2}{*}{$\clubsuit$} \\
 &  &\ &\ Test&  0.37 & 46.39 & 0.08 & 63.25 & 0.24 & 35.01 & 0.29 & 20.46 & 0.51 & 71.07 &  \\

 \hline

\multirow{8}{*}{denseNet}& \multirow{2}{*}{64} & \multirow{2}{*}{cifar10}&\ Train & 0.49 & 48.55 & 0.03 & 39.13 & 0.04 & 30.94 & 0.43 & 61.55 & 0.53 & 51.02 & \multirow{2}{*}{$\clubsuit\heartsuit\diamondsuit^*\spadesuit^*$} \\
 &  &\ &\ Test&  0.50 & 48.88 & 0.03 & 39.91 & 0.04 & 30.40 & 0.46 & 63.06 & 0.53 & 51.15 &  \\
 & \multirow{2}{*}{128} & \multirow{2}{*}{cifar10}&\ Train & 0.44 & 52.04 & 0.04 & 45.04 & 0.06 & 27.87 & 0.42 & 62.68 & 0.51 & 53.86 &  \multirow{2}{*}{$\clubsuit\heartsuit\diamondsuit^*\spadesuit^*$} \\
 &  &\ &\ Test&  0.45 & 52.37 & 0.05 & 46.20 & 0.06 & 28.16 & 0.44 & 65.07 & 0.51 & 53.70 &  \\
 & \multirow{2}{*}{256} & \multirow{2}{*}{cifar10}&\ Train & 0.46 & 47.52 & 0.09 & 42.02 & 0.12 & 27.78 & 0.37 & 71.74 & 0.50 & 49.66 & \multirow{2}{*}{$\clubsuit\heartsuit\spadesuit^*$}\\
 &  &\ &\ Test&  0.47 & 48.06 & 0.09 & 43.23 & 0.12 & 27.64 & 0.39 & 75.66 & 0.50 & 49.62 &  \\
& \multirow{2}{*}{512} & \multirow{2}{*}{cifar10}&\ Train & 0.41 & 55.20 & 0.12 & 37.63 & 0.15 & 28.04 & 0.34 & 70.05 & 0.5 & 54.92 & \multirow{2}{*}{$\clubsuit\spadesuit^*$}\\
 &  &\ &\ Test&  0.42 & 56.10 & 0.12 & 38.91 & 0.15 & 28.05 & 0.36 & 74.02 & 0.50 & 54.78 &  \\

\end{tabular}
}
\end{table}

\subsubsection{Hypotheses reviewed}
We end the presentation of our results with a review of the hypotheses posed and evidence gathered for or against each hypothesis. Out of the 8 hypotheses discussed, we found strong reason to reject hypothesis 2, weak evidence for rejection of hypothesis 4,  strong confirmatory evidence for hypotheses 1,3,7 \& 8 and weak evidence for hypothesis 5 \& 6, which would necessitate a more extensive confirmatory study design.
\begin{enumerate}
    \item We proposed that memorizing layers would have on average a lower firing rate in. This seems to be the case as long as the firing rate is evaluated on the layer responsible for memorization. Table \ref{FR-Table} and figure \ref{fig:Diff} clearly show the differences in distribution and average value for MFR in these cases.
    \item We were not able to show that the Poisson model fits better for one class of models. All models perform similar well on tests for model assumptions (see section 3.3.1). The Fano factor test even favored memorizing networks, but this might because by algorithmic issues, as discussed.
    \item We expected to observe a tendency to observe more similar behavior between random and generalizing networks. This seems to be the case for most of the models tested (\ref{FR-Table}).
    \item While we observed a tendency for a higher variability in the Firing Rate distribution (measured as CV in table \ref{FR-Table}) in some of our models, we also observed similar CV values similar to the generalizing case for other models (see figure \ref{fig:GenGap}.
    \item We did find weak evidence for a 'dilution effect' due to layer width, where the MFR of generalizing and memorizing networks move closer together (see figure \ref{fig:Diff}. This effect would have to be investigated further with a wider range of models.
    \item Similarly, we also found weak evidence for a 'dilution effect' due to layer depth, where the FR distribution of generalizing and memorizing networks are more similar, when more layers are allowed to participate in memorization (see figure \ref{FR-Table}. We investigated, that this effect is unlikely caused by continuing memorization (see figure \ref{fig:Dur}, but overall further investigations with a wider range of models would be necessary.
    \item Table \ref{FR-Table} does show strong evidence for the hypotheses, that the parameters investigated are indeed stable regardless on which data split they are calculated on. For no model did the firing rate distribution differ between test and training data. 
    \item Figure \ref{fig:Dur} shows that the models investigated do indeed exhibit clearly visible change points and stable differences in MFR during training when the models switch from generalizing to memorizing.
\end{enumerate}

\section{Conclusion}
\subsection{Discussion}
In this paper, we presented for the first time an attempt to describe the spiking behavior of artificial neural networks, trained under different learning strategies (generalization and memorization), in terms of stochastic processes.  We observed the activation within ANNs as binary spike trains and modeled them as a Poisson arrival process. This approach already  proved useful for testing hypotheses regarding the behavior of these networks during generalization and memorization. Extensive statistical tests for the key assumptions underlying Poisson process models, did show a high agreement with observed data. However, one test based on the Fano factor did provide evidence against the Poisson process fitting our experimental data and the observed processes might exhibit overall lower variability than expected by a Poisson process model. Future research should investigate and empirically compare other process models. As this paper demonstrated, correctly characterizing an appropriate model for activation patterns offers new theoretical approaches to understand learning connections systems as well as practical applications.\\
The experiments presented here show that calculated descriptive statistics of the observed process can be useful to characterize a given network, regardless of which stochastic process would be the best theoretical fit to describe spike trains produced within an ANN.  The first valuable finding was, that the distribution of spike trains observed from the known training or the unknown testing data do not differ significantly. This means that indicators describing the spike trains should be expected to be useful even without the need of an external test set.\\
In this paper, we focused on characterizing the differences between networks that learned to generalize and networks that learned to memorize their input data. For this task, we found that the calculated statistics show some promising features. The mean of the firing rate ($\lambda$) over the observed layer is consistently lower in networks trained for memorization, compared to higher firing rates in generalizing models. The width of the memorizing layer has an impact on this effect, as well as the depth of previous layers that are allowed to memorize. In the case of deeper networks, the difference between memorization and generalization did dissolve somewhat. As we only observed the last hidden layer, this results hints that memorization happens in earlier layers of the network. We also observed the firing rate of the network during training and showed a characteristic curve of the MFR during training with a quick stabilization during memorization.\\

\subsection{Future work recommendation \& Outlook}
The present study requires  more extensive research, covering additional models and datasets that would be necessary to validate and generalize our findings. We did focus on common datasets and models for image recognition, and believe that we covered a wide range from the simplest to more complex models. While other model architectures, tasks or data sets should be investigated in future studies, we currently don't have reason to assume that different data sources would change the results drastically We also limited ourselves to observing one specific layer in the network. This seems to work well, if the setup is forcing the network to memorize within this layer (a realistic setup for transfer learning), but could be expanded to the whole network. We also report results aggregated on the observed layer, using the mean and CV, while describing singular nodes could be promising as well.\\
\\
We believe that the presented methodology offers an intriguing way to look at the behavior of artificial neural networks. For example, we showed in our experiments, how the generalizing networks tend to end up with spike train statistics that are quite similar to the randomly initialized network, which isn't the case for memorizing networks. Could this be seen as an indication that generalization is 'simpler' than memorization because of the initialization? We also observe that the MFR of a memorizing network is apparently more stable than the MFR of a generalizing network, which exhibits some downward slope. Theoretical considerations like that could be addressed using the simplifying framework of stochastic processes \\
Future studies should also examine more complex models for stochastic processes. These could then be used to generate more descriptive statistics of the learning process or as input into theoretical simulation studies, to more fundamentally understand the learning dynamics of neural networks. \\
\\
Besides such theoretical questions, we also hope that the use of Poisson process parameters as indicators for memorization will find practical use cases. In this paper, we studied a rather theoretical example of forced memorization, using a randomized dataset. This is a valuable approach for theoretical insights into edge cases, but it limits us somewhat to extrapolate our findings to practical applications. Future studies should therefore look into memorization, overfitting, and other causes for generalization error 'in the wild' based on real world examples. We propose to investigate pretrained models in a transfer learning setting, to observe how indicators like the mean firing rate changes during model adaptation on small datasets. Other potential applications could be network pruning or regularization.

\bibliographystyle{unsrt}  
\bibliography{StephansBib}  

\begin{thebibliography}{10}

\bibitem{berner_modern_2021}
Julius Berner, Philipp Grohs, Gitta Kutyniok, and Philipp Petersen.
\newblock The modern mathematics of deep learning, 2021.

\bibitem{jin_quantifying_2020}
Pengzhan Jin, Lu~Lu, Yifa Tang, and George~Em Karniadakis.
\newblock Quantifying the generalization error in deep learning in terms of
  data distribution and neural network smoothness.
\newblock {\em Neural Networks}, 130:85--99, October 2020.

\bibitem{jin_how_2020}
Gaojie Jin, Xinping Yi, Liang Zhang, Lijun Zhang, Sven Schewe, and Xiaowei
  Huang.
\newblock How does {Weight} {Correlation} {Affect} the {Generalisation}
  {Ability} of {Deep} {Neural} {Networks}, October 2020.
\newblock arXiv:2010.05983 [cs].

\bibitem{laakom_within-layer_2021}
Firas Laakom, Jenni Raitoharju, Alexandros Iosifidis, and Moncef Gabbouj.
\newblock Within-layer {Diversity} {Reduces} {Generalization} {Gap}.
\newblock July 2021.
\newblock Publisher: Zenodo.

\bibitem{neyshabur_exploring_2017}
Behnam Neyshabur, Srinadh Bhojanapalli, David McAllester, and Nati Srebro.
\newblock Exploring generalization in deep learning.
\newblock {\em Advances in neural information processing systems}, 30, 2017.

\bibitem{zhang_understanding_2017}
Chiyuan Zhang, Samy Bengio, Moritz Hardt, Benjamin Recht, and Oriol Vinyals.
\newblock Understanding deep learning requires rethinking generalization,
  February 2017.
\newblock arXiv:1611.03530 [cs].

\bibitem{roberts_principles_2022}
Daniel~A. Roberts, Sho Yaida, and Boris Hanin.
\newblock {\em The {Principles} of {Deep} {Learning} {Theory}}.
\newblock May 2022.
\newblock arXiv:2106.10165 [hep-th, stat].

\bibitem{gain_abstraction_2020}
Alex Gain and Hava Siegelmann.
\newblock Abstraction mechanisms predict generalization in deep neural
  networks.
\newblock In {\em International {Conference} on {Machine} {Learning}}, pages
  3357--3366. PMLR, 2020.

\bibitem{liu_neuron_2023}
Yibing Liu, Chris~Xing Tian, Haoliang Li, Lei Ma, and Shiqi Wang.
\newblock Neuron {Activation} {Coverage}: {Rethinking} {Out}-of-distribution
  {Detection} and {Generalization}, June 2023.
\newblock arXiv:2306.02879 [cs].

\bibitem{banerjee_empirical_2019}
Chaity Banerjee, Tathagata Mukherjee, and Eduardo Pasiliao.
\newblock An {Empirical} {Study} on {Generalizations} of the {ReLU}
  {Activation} {Function}.
\newblock In {\em Proceedings of the 2019 {ACM} {Southeast} {Conference}},
  pages 164--167, Kennesaw GA USA, April 2019. ACM.

\bibitem{guiroy_improving_2022}
Simon Guiroy, Christopher Pal, Gonçalo Mordido, and Sarath Chandar.
\newblock Improving {Meta}-{Learning} {Generalization} with
  {Activation}-{Based} {Early}-{Stopping}, August 2022.
\newblock arXiv:2208.02377 [cs, stat].

\bibitem{nguyen2016multifaceted}
Anh Nguyen, Jason Yosinski, and Jeff Clune.
\newblock Multifaceted feature visualization: {Uncovering} the different types
  of features learned by each neuron in deep neural networks.
\newblock {\em arXiv preprint arXiv:1602.03616}, 2016.

\bibitem{yosinski2015understanding}
Jason Yosinski, Jeff Clune, Anh Nguyen, Thomas Fuchs, and Hod Lipson.
\newblock Understanding neural networks through deep visualization.
\newblock {\em arXiv preprint arXiv:1506.06579}, 2015.

\bibitem{selvaraju2017grad}
Ramprasaath~R Selvaraju, Michael Cogswell, Abhishek Das, Ramakrishna Vedantam,
  Devi Parikh, and Dhruv Batra.
\newblock Grad-cam: {Visual} explanations from deep networks via gradient-based
  localization.
\newblock In {\em Proceedings of the {IEEE} international conference on
  computer vision}, pages 618--626, 2017.

\bibitem{nguyen2016synthesizing}
Anh Nguyen, Alexey Dosovitskiy, Jason Yosinski, Thomas Brox, and Jeff Clune.
\newblock Synthesizing the preferred inputs for neurons in neural networks via
  deep generator networks.
\newblock {\em Advances in neural information processing systems}, 29, 2016.

\bibitem{simonyan2014deep}
K~Simonyan, A~Vedaldi, and A~Zisserman.
\newblock Deep inside convolutional networks: visualising image classification
  models and saliency maps.
\newblock In {\em Proceedings of the international conference on learning
  representations ({ICLR})}, 2014.
\newblock tex.organization: ICLR.

\bibitem{adebayo2018sanity}
Julius Adebayo, Justin Gilmer, Michael Muelly, Ian Goodfellow, Moritz Hardt,
  and Been Kim.
\newblock Sanity checks for saliency maps.
\newblock {\em Advances in neural information processing systems}, 31, 2018.

\bibitem{pizarroso2020neuralsens}
Jaime Pizarroso, José Portela, and Antonio Muñoz.
\newblock {NeuralSens}: sensitivity analysis of neural networks.
\newblock {\em arXiv preprint arXiv:2002.11423}, 2020.

\bibitem{nguyen2015deep}
Anh Nguyen, Jason Yosinski, and Jeff Clune.
\newblock Deep neural networks are easily fooled: {High} confidence predictions
  for unrecognizable images.
\newblock In {\em Proceedings of the {IEEE} conference on computer vision and
  pattern recognition}, pages 427--436, 2015.

\bibitem{blalock2020state}
Davis Blalock, Jose~Javier Gonzalez~Ortiz, Jonathan Frankle, and John Guttag.
\newblock What is the state of neural network pruning?
\newblock {\em Proceedings of machine learning and systems}, 2:129--146, 2020.

\bibitem{ye2020accelerating}
Xucheng Ye, Pengcheng Dai, Junyu Luo, Xin Guo, Yingjie Qi, Jianlei Yang, and
  Yiran Chen.
\newblock Accelerating {CNN} training by pruning activation gradients.
\newblock In {\em Computer {Vision}–{ECCV} 2020: 16th european conference,
  glasgow, {UK}, august 23–28, 2020, proceedings, part {XXV} 16}, pages
  322--338, 2020.
\newblock tex.organization: Springer.

\bibitem{anwar2017structured}
Sajid Anwar, Kyuyeon Hwang, and Wonyong Sung.
\newblock Structured pruning of deep convolutional neural networks.
\newblock {\em ACM Journal on Emerging Technologies in Computing Systems
  (JETC)}, 13(3):1--18, 2017.
\newblock Publisher: ACM New York, NY, USA.

\bibitem{zhao2022adaptive}
Kaiqi Zhao, Animesh Jain, and Ming Zhao.
\newblock Adaptive activation-based structured pruning, 2022.

\bibitem{hu2016network}
Hengyuan Hu, Rui Peng, Yu-Wing Tai, and Chi-Keung Tang.
\newblock Network trimming: {A} data-driven neuron pruning approach towards
  efficient deep architectures.
\newblock {\em arXiv preprint arXiv:1607.03250}, 2016.

\bibitem{tan2020dropnet}
Chong Min~John Tan and Mehul Motani.
\newblock Dropnet: {Reducing} neural network complexity via iterative pruning.
\newblock In {\em International conference on machine learning}, pages
  9356--9366, 2020.
\newblock tex.organization: PMLR.

\bibitem{ding2019regularizing}
Ruizhou Ding, Ting-Wu Chin, Zeye Liu, and Diana Marculescu.
\newblock Regularizing activation distribution for training binarized deep
  networks.
\newblock In {\em Proceedings of the {IEEE}/{CVF} conference on computer vision
  and pattern recognition}, pages 11408--11417, 2019.

\bibitem{joo2020regularizing}
Taejong Joo, Donggu Kang, and Byunghoon Kim.
\newblock Regularizing activations in neural networks via distribution matching
  with the {Wasserstein} metric.
\newblock {\em arXiv preprint arXiv:2002.05366}, 2020.

\bibitem{qi2019activity}
Yu~Qi, Hanwen Wang, Rui Liu, Bian Wu, Yueming Wang, and Gang Pan.
\newblock Activity-dependent neuron model for noise resistance.
\newblock {\em Neurocomputing}, 357:240--247, 2019.
\newblock Publisher: Elsevier.

\bibitem{hanin2019deep}
Boris Hanin and David Rolnick.
\newblock Deep relu networks have surprisingly few activation patterns.
\newblock {\em Advances in neural information processing systems}, 32, 2019.

\bibitem{merity2017revisiting}
Stephen Merity, Bryan McCann, and Richard Socher.
\newblock Revisiting activation regularization for language {RNNs}.
\newblock {\em arXiv e-prints}, pages arXiv--1708, 2017.

\bibitem{shadlen1994noise}
Michael~N Shadlen and William~T Newsome.
\newblock Noise, neural codes and cortical organization.
\newblock {\em Current opinion in neurobiology}, 4(4):569--579, 1994.
\newblock Publisher: Elsevier.

\bibitem{Softky334}
WR~Softky and C~Koch.
\newblock The highly irregular firing of cortical cells is inconsistent with
  temporal integration of random {EPSPs}.
\newblock {\em Journal of Neuroscience}, 13(1):334--350, 1993.
\newblock Publisher: Society for Neuroscience tex.eprint:
  https://www.jneurosci.org/content/13/1/334.full.pdf.

\bibitem{deger2009poisson}
Moritz Deger, Stefano Cardanobile, Moritz Helias, and Stefan Rotter.
\newblock The {Poisson} process with dead time captures important statistical
  features of neural activity.
\newblock {\em BMC Neuroscience}, 10(Suppl 1):P110, 2009.
\newblock Publisher: BioMed Central London.

\bibitem{reynaud2013spike}
Patricia Reynaud-Bouret, Christine Tuleau-Malot, Vincent Rivoirard, Franck
  Grammont, and {others}.
\newblock Spike trains as (in) homogeneous {Poisson} processes or {Hawkes}
  processes: non-parametric adaptive estimation and goodness-of-fit tests.
\newblock {\em Journal of Mathematical Neuroscience}, 39(8):32--33, 2013.

\bibitem{PhysRevE.73.022901}
Benjamin Lindner.
\newblock Superposition of many independent spike trains is generally not a
  {Poisson} process.
\newblock {\em Physical Review E: Statistical Physics, Plasmas, Fluids, and
  Related Interdisciplinary Topics}, 73(2):022901, February 2006.
\newblock Number of pages: 4 Publisher: American Physical Society.

\bibitem{Kass2014}
Robert~E. Kass, Uri~T. Eden, and Emery~N. Brown.
\newblock Point processes.
\newblock In {\em Analysis of neural data}, pages 563--603. Springer New York,
  New York, NY, 2014.

\bibitem{kramer2016case}
Mark~A Kramer and Uri~T Eden.
\newblock {\em Case studies in neural data analysis: a guide for the practicing
  neuroscientist}.
\newblock MIT Press, 2016.

\bibitem{brown2001stochastic}
Bradley~D Brown and Howard~C Card.
\newblock Stochastic neural computation. {I}. {Computational} elements.
\newblock {\em IEEE Transactions on computers}, 50(9):891--905, 2001.
\newblock Publisher: IEEE.

\bibitem{963787}
H.C. Card.
\newblock Compound binomial processes in neural integration.
\newblock {\em IEEE Transactions on Neural Networks}, 12(6):1505--1512, 2001.

\bibitem{1058080}
H.C. Card and D.K. McNeill.
\newblock Gaussian activation functions using {Markov} chains.
\newblock {\em IEEE Transactions on Neural Networks}, 13(6):1465--1471, 2002.

\bibitem{CARD2001173}
Howard~C. Card.
\newblock Dynamics of stochastic artificial neurons.
\newblock {\em Neurocomputing}, 41(1):173--182, 2001.

\bibitem{cowan1990stochastic}
Jack~D Cowan.
\newblock Stochastic neurodynamics.
\newblock {\em Advances in neural information processing systems}, 3, 1990.

\bibitem{card1998doubly}
Howard~C Card.
\newblock Doubly stochastic {Poisson} processes in artificial neural learning.
\newblock {\em IEEE transactions on neural networks}, 9(1):229--231, 1998.
\newblock Publisher: IEEE.

\bibitem{doi:10.1142/S0129065701000552}
HOWARD~C. CARD.
\newblock {STOCHASTIC} {RADIAL} {BASIS} {FUNCTIONS}.
\newblock {\em International Journal of Neural Systems}, 11(02):203--210, 2001.
\newblock tex.eprint: https://doi.org/10.1142/S0129065701000552.

\bibitem{YANG201887}
Fenglian Yang, Liang Yan, and Leevan Ling.
\newblock Doubly stochastic radial basis function methods.
\newblock {\em Journal of Computational Physics}, 363:87--97, 2018.

\bibitem{pregowska_signal_2021}
Agnieszka Pregowska.
\newblock Signal {Fluctuations} and the {Information} {Transmission} {Rates} in
  {Binary} {Communication} {Channels}.
\newblock {\em Entropy}, 23(1), 2021.

\bibitem{PhysRevA.44.2718}
Tom~M. Heskes and Bert Kappen.
\newblock Learning processes in neural networks.
\newblock {\em Physical Review A: Atomic, Molecular, and Optical Physics},
  44(4):2718--2726, August 1991.
\newblock Number of pages: 0 Publisher: American Physical Society.

\bibitem{goltsev2010stochastic}
AV~Goltsev, FV~De~Abreu, SN~Dorogovtsev, and JFF Mendes.
\newblock Stochastic cellular automata model of neural networks.
\newblock {\em Physical Review E}, 81(6):061921, 2010.
\newblock Publisher: APS.

\bibitem{keane2001impulses}
John~F Keane and Les~E Atlas.
\newblock Impulses and stochastic arithmetic for signal processing.
\newblock In {\em 2001 {IEEE} international conference on acoustics, speech,
  and signal processing. {Proceedings} (cat. {No}. {01CH37221})}, volume~2,
  pages 1257--1260, 2001.
\newblock tex.organization: IEEE.

\bibitem{ma2012high}
Chengguang Ma, Shunan Zhong, and Hua Dang.
\newblock High fault tolerant image processing system based on stochastic
  computing.
\newblock In {\em 2012 international conference on computer science and service
  system}, pages 1587--1590, 2012.
\newblock tex.organization: IEEE.

\bibitem{661204}
B.~Coker, M.~Pradier, and F.~Doshi-Velez.
\newblock Towards expressive priors for bayesian neural networks: {Poisson}
  process radial basis function networks.
\newblock In {\em proceedings at the conference on uncertainty in artificial
  intelligence ({UAI})}, volume~1, pages 1--37, 2019.

\bibitem{hanin2021random}
Boris Hanin.
\newblock Random neural networks in the infinite width limit as {Gaussian}
  processes.
\newblock {\em arXiv preprint arXiv:2107.01562}, 2021.

\bibitem{PhysRevD.103.116023}
Lijia Jiang, Lingxiao Wang, and Kai Zhou.
\newblock Deep learning stochastic processes with {QCD} phase transition.
\newblock {\em Physical Review D: Particles and Fields}, 103(11):116023, June
  2021.
\newblock Number of pages: 8 Publisher: American Physical Society.

\bibitem{Lee_Zame_Yoon_vanderSchaar_2018}
Changhee Lee, William Zame, Jinsung Yoon, and Mihaela van~der Schaar.
\newblock {DeepHit}: {A} deep learning approach to survival analysis with
  competing risks.
\newblock {\em Proceedings of the AAAI Conference on Artificial Intelligence},
  32(1), April 2018.

\bibitem{NEURIPS2021_f19c44d0}
Emile Mathieu, Adam Foster, and Yee Teh.
\newblock On contrastive representations of stochastic processes.
\newblock In M.~Ranzato, A.~Beygelzimer, Y.~Dauphin, P.S. Liang, and J.~Wortman
  Vaughan, editors, {\em Advances in neural information processing systems},
  volume~34, pages 28823--28835. Curran Associates, Inc., 2021.

\bibitem{cinlar2013introduction}
Erhan Cinlar.
\newblock {\em Introduction to stochastic processes}.
\newblock Courier Corporation, 2013.

\bibitem{nelson2010stochastic}
Barry~L Nelson.
\newblock {\em Stochastic modeling: analysis \& simulation}.
\newblock Courier Corporation, 2010.

\bibitem{GABBIANI2017335}
Fabrizio Gabbiani and Steven~James Cox.
\newblock Chapter 18 - stochastic processes.
\newblock In Fabrizio Gabbiani and Steven~James Cox, editors, {\em Mathematics
  for neuroscientists (second edition)}, pages 335--349. Academic Press, San
  Diego, second edition edition, 2017.

\bibitem{maimon_beyond_2009}
Gaby Maimon and John~A. Assad.
\newblock Beyond {Poisson}: {Increased} {Spike}-{Time} {Regularity} across
  {Primate} {Parietal} {Cortex}.
\newblock {\em Neuron}, 62(3):426--440, May 2009.

\bibitem{stella_comparing_2021}
Alessandra Stella, Peter Bouss, Günther Palm, and Sonja Grün.
\newblock Comparing surrogates to evaluate precisely timed higher-order spike
  correlations, December 2021.

\bibitem{berry_structure_1997}
Michael~J. Berry, David~K. Warland, and Markus Meister.
\newblock The structure and precision of retinal spike trains.
\newblock {\em Proceedings of the National Academy of Sciences},
  94(10):5411--5416, May 1997.

\bibitem{ramezan_multiscale_2014}
Reza Ramezan, Paul Marriott, and Shojaeddin Chenouri.
\newblock Multiscale analysis of neural spike trains.
\newblock {\em Statistics in Medicine}, 33(2):238--256, January 2014.

\bibitem{naud_improved_2011}
Richard Naud, Felipe Gerhard, Skander Mensi, and Wulfram Gerstner.
\newblock Improved {Similarity} {Measures} for {Small} {Sets} of {Spike}
  {Trains}.
\newblock {\em Neural Computation}, 23(12):3016--3069, December 2011.

\bibitem{williams_point_2020}
Alex~H. Williams, Anthony Degleris, Yixin Wang, and Scott~W. Linderman.
\newblock Point process models for sequence detection in high-dimensional
  neural spike trains.
\newblock {\em Advances in neural information processing systems},
  33:14350--14361, December 2020.

\bibitem{zhang_understanding_2021}
Chiyuan Zhang, Samy Bengio, Moritz Hardt, Benjamin Recht, and Oriol Vinyals.
\newblock Understanding deep learning (still) requires rethinking
  generalization.
\newblock {\em Communications of the ACM}, 64(3):107--115, 2021.
\newblock Publisher: ACM New York, NY, USA.

\bibitem{stephenson_geometry_2021}
Cory Stephenson, Suchismita Padhy, Abhinav Ganesh, Yue Hui, Hanlin Tang, and
  SueYeon Chung.
\newblock On the geometry of generalization and memorization in deep neural
  networks.
\newblock {\em arXiv preprint arXiv:2105.14602}, 2021.

\bibitem{mo_towards_2019}
Fan Mo, Ali~Shahin Shamsabadi, Kleomenis Katevas, Andrea Cavallaro, and Hamed
  Haddadi.
\newblock Towards characterizing and limiting information exposure in {DNN}
  layers.
\newblock {\em arXiv preprint arXiv:1907.06034}, 2019.

\bibitem{arpit_closer_2017}
Devansh Arpit, Stanis{\textbackslash}law Jastrzębski, Nicolas Ballas, David
  Krueger, Emmanuel Bengio, Maxinder~S Kanwal, Tegan Maharaj, Asja Fischer,
  Aaron Courville, Yoshua Bengio, and {others}.
\newblock A closer look at memorization in deep networks.
\newblock In {\em International conference on machine learning}, pages
  233--242. PMLR, 2017.

\bibitem{cohen_dnn_2018}
Gilad Cohen, Guillermo Sapiro, and Raja Giryes.
\newblock {DNN} or k-{NN}: {That} is the {Generalize} vs. {Memorize}
  {Question}.
\newblock {\em arXiv preprint arXiv:1805.06822}, 2018.

\bibitem{wongso_using_2023}
Shelvia Wongso, Rohan Ghosh, and Mehul Motani.
\newblock Using {Sliced} {Mutual} {Information} to {Study} {Memorization} and
  {Generalization} in {Deep} {Neural} {Networks}.
\newblock In Francisco Ruiz, Jennifer Dy, and Jan-Willem van~de Meent, editors,
  {\em Proceedings of {The} 26th {International} {Conference} on {Artificial}
  {Intelligence} and {Statistics}}, volume 206 of {\em Proceedings of {Machine}
  {Learning} {Research}}, pages 11608--11629. PMLR, April 2023.

\bibitem{chollet_keras_2015}
François Chollet and {others}.
\newblock Keras, 2015.

\bibitem{huang_densely_2017}
Gao Huang, Zhuang Liu, Laurens Van Der~Maaten, and Kilian~Q. Weinberger.
\newblock Densely {Connected} {Convolutional} {Networks}.
\newblock In {\em 2017 {IEEE} {Conference} on {Computer} {Vision} and {Pattern}
  {Recognition} ({CVPR})}, pages 2261--2269, Honolulu, HI, July 2017. IEEE.

\bibitem{glorot_understanding_2010}
Xavier Glorot and Yoshua Bengio.
\newblock Understanding the difficulty of training deep feedforward neural
  networks.
\newblock In {\em Proceedings of the thirteenth international conference on
  artificial intelligence and statistics}, pages 249--256. JMLR Workshop and
  Conference Proceedings, 2010.

\bibitem{eden_drawing_2010}
Uri~T Eden and Mark~A Kramer.
\newblock Drawing inferences from {Fano} factor calculations.
\newblock {\em Journal of neuroscience methods}, 190(1):149--152, 2010.
\newblock Publisher: Elsevier.

\bibitem{rajdl_fano_2020}
Kamil Rajdl, Petr Lansky, and Lubomir Kostal.
\newblock Fano {Factor}: {A} {Potentially} {Useful} {Information}.
\newblock {\em Frontiers in Computational Neuroscience}, 14:569049, November
  2020.

\bibitem{cox_renewal_1962}
D.R. Cox.
\newblock {\em Renewal {Theory} by {D}.{R}. {Cox}}.
\newblock Methuen's monographs on applied probability and statistics. Methuen;
  New York, Wiley, 1962.

\bibitem{ljung_measure_1978}
G.~M. Ljung and G.~E.~P. Box.
\newblock On a measure of lack of fit in time series models.
\newblock {\em Biometrika}, 65(2):297--303, August 1978.

\bibitem{heard_choosing_2018}
N~A Heard and P~Rubin-Delanchy.
\newblock Choosing between methods of combining \$p\$-values.
\newblock {\em Biometrika}, 105(1):239--246, March 2018.

\bibitem{seabold_statsmodels_2010}
Skipper Seabold and Josef Perktold.
\newblock statsmodels: {Econometric} and statistical modeling with python.
\newblock In {\em 9th {Python} in {Science} {Conference}}, 2010.

\bibitem{messer_multiple_2014}
Michael Messer, Marietta Kirchner, Julia Schiemann, Jochen Roeper, Ralph
  Neininger, and Gaby Schneider.
\newblock A multiple filter test for the detection of rate changes in renewal
  processes with varying variance.
\newblock {\em The Annals of Applied Statistics}, 8(4), December 2014.

\end{thebibliography}
\newpage
\section*{Appendix}

\subsection*{Software Packages}

\begin{table}[h]
\centering
\caption{Software used for the experiments in this paper}
\begin{tabular}{ |c|c|c| } 
 \hline
 \textbf{Software} & \textbf{Version} & \textbf{Link} \\ \hline 
 Python & 3.8.5 & \url{https://www.python.org/}\\
 Matplotlib & 3.3.2 & \url{https://matplotlib.org/} \\
 Seaborn & 0.11.0 & \url{https://seaborn.pydata.org/} \\
 Tensorflow & 2.9.1 & \url{https://www.tensorflow.org/}\\
 Pandas & 1.2.4 & \url{https://pandas.pydata.org/} \\
 Numpy & 1.21.2 & \url{https://numpy.org/} \\
 Elephant & 0.13.0 & \url{https://elephant.readthedocs.io/en/latest/} \\
 Neocore & 0.5.6 & \url{https://neo-python-core.readthedocs.io/}\\
 Quantities & 0.14.1 & \url{https://python-quantities.readthedocs.io/}\\
 Statsmodel & 0.14.0 & \url{www.statsmodels.org}\\
 Scipy.Stats & 1.11.1 & \url{https://scipy.org/}\\
 Ptitprince & 0.2.6 & \url{https://github.com/pog87/PtitPrince}\\
 \hline
\end{tabular}

\label{table:APPsoftware}
\end{table}


\subsection*{Workstation}

\begin{table}[h]
\caption{Hardware used for the experiments in this paper}
\centering
\begin{tabular}{ |c|c| } 
 \hline
 & \textbf{Type} \\ \hline 
 OS & Windows 10\\
 CPU & AMD Ryzen 5 3600 \\
 GPU & NVIDIA GeForce RTX 2060\\
 CUDA Capability & 7.5 \\
 RAM & 64 GB \\
 SSD & 1 TB \\
 \hline
\end{tabular}

\label{table:APPworkstation}
\end{table}

\subsection*{Additional Plots}

\subsection*{Additional Tables}
\begin{table}[htp]
\centering
\caption{\label{table:mf} Mean Fano Factor of every network}
\resizebox{\textwidth}{!}{
\begin{tabular}{c|c|c|c|c|c|c|c|c}
\hline
\multirow{2}{*}{Model} & \multirow{2}{*}{Width} & \multirow{2}{*}{Data} & \multirow{2}{*}{Data split} & Generalizing & MemRetrainLast & MemRandomLast & MemRetrainAll & Random\\
&&&&  Mean FF  & Mean FF & Mean FF & Mean FF & Mean FF\\
\hline
\multirow{16}{*}{Dense} & \multirow{4}{*}{64} & \multirow{2}{*}{MNIST}&\ Train & 0.38 & 0.97* & 0.87 & - & 0.55 \\
 &  &\ &\ Test& 0.38 & 0.95* & 0.85 & - & 0.55 \\
 &  &\ \multirow{2}{*}{FashionMNIST}&\ Train & 0.49 & 097* & 0.91 & - & 0.48 \\
 &  &\  &\  Test & 0.51 & 0.95* & 0.91* & - & 0.48 \\
 & \multirow{4}{*}{128} & \multirow{2}{*}{MNIST}&\ Train & 0.40 & 0.90 & 0.81 & - & 0.52 \\
 &  &\ &\ Test& 0.40 & 0.88* & 0.80 & - & 0.51 \\
 &  &\ \multirow{2}{*}{FashionMNIST}&\ Train & 0.51 &  0.95* & 0.88 & - & 0.53 \\
 &  &\  &\  Test & 0.52 &  0.95* & 0.88* & - & 0.52\\
  & \multirow{4}{*}{256} & \multirow{2}{*}{MNIST}&\ Train & 0.40 & 0.85 & 0.74 & - & 0.50 \\
 &  &\ &\ Test& 0.42 & 0.86 & 0.73 & - & 0.49 \\
 &  &\ \multirow{2}{*}{FashionMNIST}&\ Train & 0.55  & 0.92 & 083 & - & 0.52\\
 &  &\  &\  Test & 0.55  & 0.93* & 0.84 & - & 0.52\\
  & \multirow{4}{*}{512} & \multirow{2}{*}{MNIST}&\ Train & 0.42 & 0.82 & 0.71 & -& 0.49 \\
 &  &\ &\ Test& 0.43 & 0.81 & 0.71 & - & 0.49 \\
 &  &\ \multirow{2}{*}{FashionMNIST}&\ Train & 0.56 & 0.90 & 0.81 & - & 0.50   \\
 &  &\  &\  Test & 0.56 & 0.89 & 0.81 & - & 0.49\\
 \hline
 \multirow{16}{*}{CNN} & \multirow{4}{*}{64} & \multirow{2}{*}{MNIST}&\ Train & 0.52 & 0.99* & 0.76 & 0.71 & 0.51 \\
 &  &\ &\ Test& 0.52 & 0.97* & 0.73 & 0.71 & 0.52 \\
 &  &\ \multirow{2}{*}{FashionMNIST}&\ Train & 0.54 &  0.95* & 0.81 & 0.73 & 0.47\\
 &  &\  &\  Test & 0.53 &  0.95* & 0.82 & 0.71 & 0.47\\
 & \multirow{4}{*}{128} & \multirow{2}{*}{MNIST}&\ Train & 0.53 & 0.98* & 0.70 & 0.71 & 0.48 \\
 &  &\ &\ Test& 0.54 & 0.97* & 0.69 & 0.71 & 0.49 \\
 &  &\ \multirow{2}{*}{FashionMNIST}&\ Train & 0.56 &  0.94 & 0.78 & 0.70 & 0.50 \\
 &  &\  &\  Test & 0.55 &  0.94* & 0.77 & 0.70 & 0.50\\
  & \multirow{4}{*}{256} & \multirow{2}{*}{MNIST}&\ Train & 0.57 & 0.98* & 0.66 & 0.71 & 0.49 \\
 &  &\ &\ Test& 0.58 & 0.98* & 0.65 & 0.70 & 0.49 \\
 &  &\ \multirow{2}{*}{FashionMNIST}&\ Train & 0.59  & 0.92 & 0.68 & 0.69& 0.49\\
 &  &\  &\  Test & 0.60 &  0.91* & 0.68 & 0.68 & 0.48\\
  & \multirow{4}{*}{512} & \multirow{2}{*}{MNIST}&\ Train & 0.58 & 0.98* & 0.69 & 0.72 & 0.49 \\
 &  &\ &\ Test& 0.59 & 0.98* & 0.67 & 0.71 & 0.49 \\
 &  &\ \multirow{2}{*}{FashionMNIST}&\ Train & 0.63 & 0.92 & 0.76 & 0.71 & 0.49\\
 &  &\  &\  Test & 0.62 &  0.91 & 0.76 & 0.71 & 0.49\\
 \hline

 \multirow{8}{*}{DenseNet} & \multirow{2}{*}{64} & \multirow{2}{*}{cifar10}&\ Train & 0.53 & 0.97* & 0.96* & 0.60 & 0.47 \\
 &  &\ &\ Test& 0.48 & 0.97* & 095* & 0.52 & 0.46 \\
 & \multirow{2}{*}{128} & \multirow{2}{*}{cifar10}&\ Train & 0.56 & 0.94* & 0.93* & 0.61 & 0.50 \\
 &  &\ &\ Test& 0.52 & 0.93* & 0.92* & 0.55 & 0.49 \\
  & \multirow{2}{*}{256} & \multirow{2}{*}{cifar10}&\ Train & 0.54 & 0.92 & 0.88 & 0.66 & 0.50 \\
 &  &\ &\ Test& 0.54 & 0.91 & 0.88 & 0.62 & 0.49 \\
  & \multirow{2}{*}{512} & \multirow{2}{*}{cifar10}&\ Train & 0.59 & 0.89 & 0.85 & 0.69 & 0.47 \\
 &  &\ &\ Test& 0.57 & 0.88 & 0.84 & 0.63 & 0.46 \\
 \hline

\end{tabular}
}

\end{table}

\end{document}